\documentclass[lettersize,journal]{IEEEtran}
\usepackage{amsmath,amsfonts}
\usepackage{algorithmic}
\usepackage{algorithm}
\usepackage{array}
\usepackage{textcomp}
\usepackage{stfloats}
\usepackage{url}
\usepackage{verbatim}
\usepackage{graphicx}
\usepackage{cite}
\usepackage{indentfirst}
\usepackage{multirow}
\usepackage{hhline}
\usepackage{epstopdf}
\usepackage{booktabs}
\usepackage{threeparttable}
\usepackage{gensymb}
\usepackage{stmaryrd} 
\usepackage{xcolor}
\usepackage{authblk}
\usepackage{epsfig}
\usepackage{graphicx}
\usepackage{amssymb}
\usepackage{color}

\usepackage{amsfonts}
\usepackage{amsthm}
\usepackage{dsfont}
\usepackage{hyperref}
\usepackage{breqn}
\usepackage{diagbox}
\usepackage{pdfpages}
\usepackage{fancyref}
\usepackage{subfigure}
\usepackage[normalem]{ulem}
\definecolor{mygreen}{RGB}{146,208,80}
\definecolor{myblue}{RGB}{0,112,192}

\begin{document}

\title{Learning Kernel-Modulated Neural Representation for Efficient Light Field Compression}

\author{Jinglei~Shi, Yihong~Xu, Christine~Guillemot~\IEEEmembership{Fellow,~IEEE}}

% The paper headers
\markboth{Journal of \LaTeX\ Class Files,~Vol.~14, No.~8, August~2021}%
{Shell \MakeLowercase{\textit{et al.}}: A Sample Article Using IEEEtran.cls for IEEE Journals}

\maketitle

\begin{abstract}
Light field is a type of image data that captures the 3D scene information by recording light rays emitted from a scene at various orientations. It offers a more immersive perception than classic 2D images but at the cost of huge data volume. In this paper, we draw inspiration from the visual characteristics of Sub-Aperture Images (SAIs) of light field and design a compact neural network representation for the light field compression task. The network backbone takes randomly initialized noise as input and is supervised on the SAIs of the target light field. It is composed of two types of complementary kernels: descriptive kernels (\textit{descriptors}) that store scene description information learned during training, and modulatory kernels (\textit{modulators}) that control the rendering of different SAIs from the queried perspectives. To further enhance compactness of the network meanwhile retain high quality of the decoded light field, we accordingly introduce modulator allocation and kernel tensor decomposition mechanisms, followed by non-uniform quantization and lossless entropy coding techniques, to finally form an efficient compression pipeline. Extensive experiments demonstrate that our method outperforms other state-of-the-art (SOTA) methods by a significant margin in the light field compression task. Moreover, after aligning descriptors, the modulators learned from one light field can be transferred to new light fields for rendering dense views, indicating a potential solution for view synthesis task.
\end{abstract}


\section{Introduction}
\IEEEPARstart{L}{ight} fields \cite{levoy1996light,gortler1996lumigraph} record both the intensity and direction of light rays emitted by a scene in 3D space. The additional angular information provides users with a more immersive experience than classic 2D images when navigating within the captured scene, and powers a series of computer vision tasks such as depth estimation~\cite{shi2019depth,wang2022occlusion}, super-resolution~\cite{wang2022detail,wang2022disentangling}, instance segmentation~\cite{xu2015transcut}, salient object detection~\cite{jing2021occlusion}.
Although the spatio-angular information of light fields offers numerous benefits for various applications, it also introduces a significant challenge in terms of data volume. The inherent redundancy in light fields results in large storage requirements, increased transmission bandwidth, and demands on display hardware. Therefore, a crucial aspect in advancing light field imaging techniques towards practical scenarios is the development of effective compression solutions.

Early compression methods~\cite{lucas2014locally,Conti2016Bi,monteiro2016light} primarily concentrated on directly compressing the lenslet image obtained from plenoptic cameras with the help of HEVC-intra coding framework. However, these intra-coding-based approaches has demonstrated limited performance. 
In contrast, more general solutions~\cite{dai2015lenselet,liu2016pseudo,Ahmad2017,viola2017comparison,li2017pseudo} have employed video compression standards (in particular HEVC) to tackle the set of light field views as a pesudo video sequence, which enables exploration of temporal correlations among frames.
Besides classical video codecs, more advanced learning-based video compression solutions~\cite{lu2019dvc,yang2020Learning,yang2021learning,yang2021perceptual} can also be applied in the light field compression context.
Methods based on view synthesis technique have also been proposed in previous studies~\cite{jiang2017hot3D,huang2020low,shi2022deep} for the goal of compression, where the encoder side focuses on the compression of a subset of SAIs, and the decoders reconstruct the full light field from the received subset by applying view rendering methods. 
In~\cite{Zhao2017}, the authors utilize Matching Pursuit to perform a linear approximation, enabling disparity-based view prediction.
In~\cite{Astola2018WaSP}, a depth-based light field codec called Warping and Sparse Prediction (WaSP) was introduced, and its enhanced version has later been incorporated as the 4D-Prediction mode in the JPEG Pleno light field coding standard. The WaSP codec relies on depth-based warping and merging of warped reference views, forming the primary prediction stage.
Transformation is another effective tool to reduce light field redundancy. A 4D-Transform mode named Multidimensional Light field Encoder (MuLE)~\cite{mule} has been adopted in JPEG Pleno, where the 4D redundancy of light fields is exploited by applying a 4D-DCT transform to 4D spatio-angular blocks. The authors in~\cite{Rizkallah2020GeometryAwareGT} propose a graph-transform-based light field compression method tailored by the scene geometry. Other transforms such as mixture of expert~\cite{Verhack2020SME}, homography-based low rank approximation~\cite{Jiang2019LR} or shearlet transform~\cite{ahmad2020shearlet} show their superiority for narrow-baseline data, but suffer from performance degradation when the light fields' baseline increases.

The emergence of Neural Radiance Field (NeRF) \cite{mildenhall2020nerf} has ushered in a new era of employing implicit neural networks to represent scenes for diverse applications~\cite{bemana2020x, mildenhall2022nerf, wang2023jaws, qiu2023looking}. NeRF utilizes Multi-Layer Perceptrons (MLPs) to establish mappings between 5D coordinates (position and orientation) of light rays and color as well as density for the volumetric rendering process. Such an Implicit Neural Representation (INR) has also brought about a fresh perspective of using network weights to represent light fields for compression task. Authors in~\cite{shi2023light} propose to train a NeRF with low-rank constraint in ADMM optimization framework, followed by distillation and quantization operations, to finally obtain a compact representation of light fields. Implicit neural network can also serve as a prior in the compression context, authors in~\cite{jiang2022untrained} propose a two-stage workflow that uses a GRU to encode transient information between SAIs into latent vectors, which are then processed by a generator to retrieve blocks of light field views. Let us note that INR-based methods have created a link between the problems of light field compression and network compression. Methods like pruning~\cite{zhang2018systematic,gao2021network,wang2021convolutional}, tensor rank optimization~\cite{yin2021towards,qiu2018dcfnet}, quantization~\cite{jacob2018quantization,fan2020training} that address the compactness of deep models are therefore applicable for light field compression.

Light fields are captured by specially designed equipment \cite{adelson1992single,ng2005light,wilburn2005high} after a single exposure. On one hand, all SAIs exhibit similar visual content of the scene, but on the other hand, each SAI has its unique visual content that is only observed from corresponding perspective due to the parallax and specularity. In this paper, we propose a novel network design that draws inspiration from the above visual characteristics of light fields to address the problem of compression. The network is composed of the shared descriptive kernels (\textit{descriptors}) and individual modulatory kernels (\textit{modulators}): the descriptors will be repeatedly employed when rendering different SAIs, which mimic the fact SAIs own similar visual content. To ensure that each SAI has visual content specifically observed from the corresponding perspective, the rendering process will be guided by modulators, and each SAI corresponds to an individual set of modulators.
The network backbone takes uniformly initialized noise as input and is supervised by a randomly selected SAI stream. In the end of each training iteration, modulators of the current SAI are switched when the ground-truth SAI changes. The light field to be compressed is ultimately represented implicitly by both descriptors and modulators, where descriptors account for the majority of the network parameters for storing scene information, and minor modulators control the rendering of the desired SAI.

The essence of applying INR-based method to compression lies in achieving a delicate balance between model compactness and decoding quality, i.e. utilizing the minimum number of network parameters to represent a light field while preserving the highest possible quality of the decoded views. To address this challenge, we further propose the \textit{modulator allocation} and \textit{kernel tensor decomposition} mechanisms. The modulator allocation mechanism effectively mitigates parameter explosion, especially in scenarios where the target light field has a high angular resolution. Additionally, the kernel tensor decomposition, as a widely-used network compression technique, decomposes both high-dimensional descriptors and modulators into the product of the low-dimensional components. This decomposition strategy aims to reduce the overall parameter count while preserving the reconstruction accuracy. In our efforts towards network compactness, we also adopt a quantization-aware training strategy~\cite{jacob2018quantization} which reduces the number of bits required for each weight and detains quantization errors.

In order to validate the effectiveness of our proposed method, we quantitatively and qualitatively evaluate our method and compare it with several representative state-of-the-art (SOTA) methods tailored for light field compression, including video compression-based methods HEVC-Lozange~\cite{HEVCstandard,rizkallah2016impact}, HLVC~\cite{yang2020Learning} and RLVC~\cite{yang2021learning}, 4D-Prediction mode of the coding standard JPEG-Pleno~\cite{jpegpleno} as well as the most recent INR-based schemes DDLF~\cite{jiang2022untrained} and QDLR-NeRF~\cite{shi2023light}. Experimental results show that our method outperforms others by a large margin and yields better visual reconstruction quality. Moreover, we carried out comprehensive comparison with other two INR-based methods (DDLF and QDLR-NeRF) in terms of encoding and decoding complexity, memory consumption and generality on different types of light field, proving the superiority of our method in the context of compression. 
Besides performance gains for the task of compression, another advantage of our proposed method is that the modulators learned from one light field can be applied to the new light fields for synthesizing dense views after aligning descriptors. It not only verifies the functionality of two types of kernels, but also implies a potential view synthesis philosophy through kernel transfer.

To summarize, the contributions of our work are as follows:

\begin{itemize}
\item We propose a novel implicit representation format for light field, which is composed of the complementary \textit{descriptors} \& \textit{modulators} to respectively store scene information and control the rendering of different SAIs.
\item By introducing \textit{modulator allocation} and \textit{kernel tensor decomposition} mechanisms, the network can effectively avoid parameter explosion when light fields have high angular resolution, and reaches a better balance between the model compactness and decoding quality.
\item We carried out extensive experiments to show that our method outperforms other SOTA methods both quantitatively and qualitatively for the task of compression. It shows better generalization on different types of data and superior abilities in terms of complexity and resource consumption than other INR-based methods.
\item We further demonstrate that the learned modulators can be transferred to the new light fields, helping to generating dense views of the new light fields, implying a potentially novel view synthesis philosophy as well.

\end{itemize}

\section{Methodology}
\begin{figure*}[!htbp]
    \centering
  \includegraphics[width=0.99\linewidth]{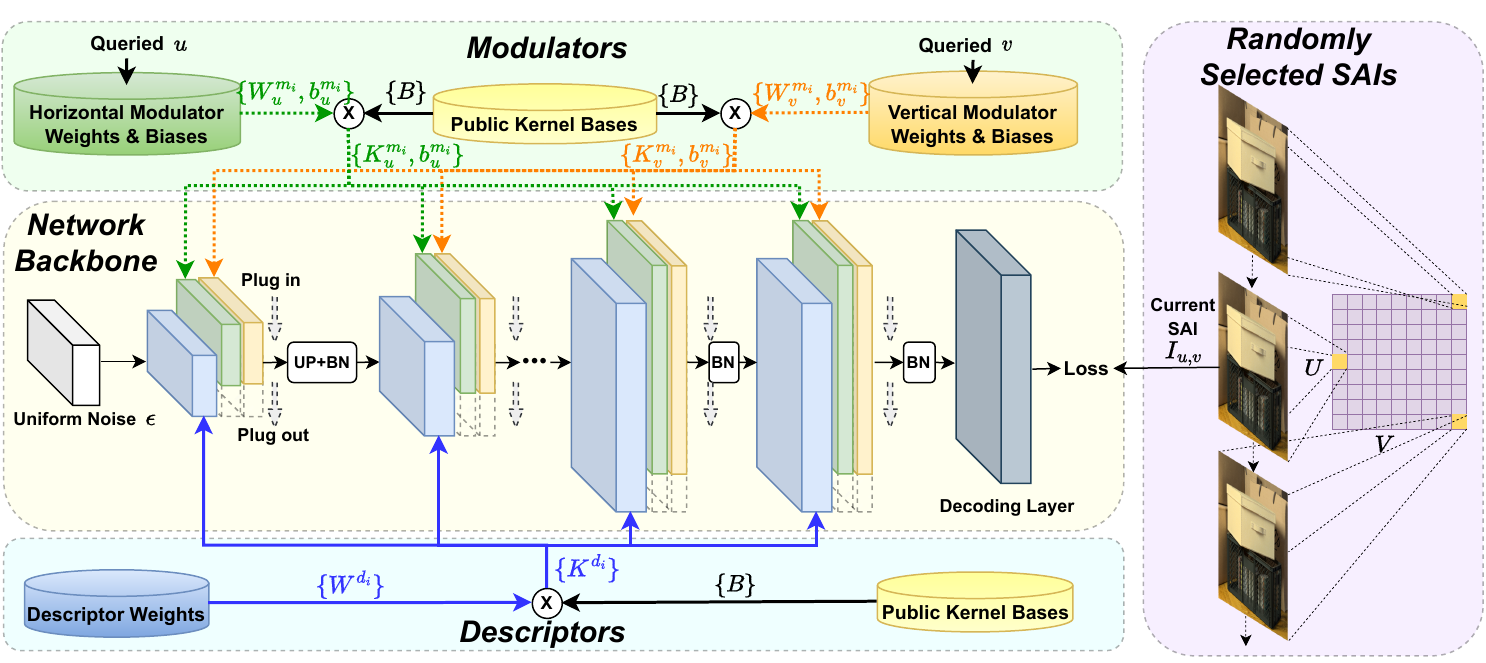}
  \caption{Network overview: the proposed \textbf{\textit{network backbone}} takes uniform noise $\epsilon$ as input and is supervised by \textbf{\textit{randomly selected SAIs}} of the target light field. It is composed of the shared \textbf{\textit{descriptors}} for storing scene information, and switchable \textbf{\textit{modulators}} for guiding the rendering of views. To further improve the compactness of the network, the modulators are allocated along the horizontal (green) and vertical (orange) directions, and both descriptors and modulators are decomposed into the production of public bases (yellow cylinder) and corresponding weights (blue, green and orange cylinders). In the end of training procedure, a light field can be implicitly represented by the shared bases $\{B\}$, descriptor weights $\{W^{d_{i}}\}$ and modulator weights and biases $\{K^{m_i}_u, K^{m_i}_v, b^{m_i}_u, b^{m_i}_v\}$.}
  \label{fig:workflow}
\end{figure*}

\subsection{Notations and network backbone}
\label{sec:overview}
We represent a light field with a 4D function $L(x,y,u,v)$ following the two-parallel-plane parameterization introduced in \cite{levoy1996light,gortler1996lumigraph}, where $(x,y) \in \llbracket 1;X \rrbracket \times \llbracket 1;Y \rrbracket$ and $(u,v) \in \llbracket 1;U \rrbracket \times \llbracket 1;V \rrbracket$ are respectively the spatial and angular coordinates. The SAI located at angular position $(u,v)$ is denoted as $I_{u,v}$ throughout the rest of the paper for simplicity. 

The goal of our work is to discover a compact neural representation for light fields that contains a limited number of parameters while still being able to retrieve high quality views. In a previous study~\cite{heckel_deep_2018}, a deep convolutional decoder was proposed to fit images for tasks such as compression, inpainting, or denoising. However, this approach was specifically designed for single RGB image and is not directly applicable to light fields. Although a straightforward solution would be to train the same number of deep decoders as there are SAIs in the target light field, this would be time-consuming and requires a large number of parameters, hence is not suitable for light field compression.
In the subsequent work~\cite{jiang2022untrained}, an advancement was made by cascading a Gated Recurrent Unit (GRU)~\cite{cho2014learning} architecture and a deep decoder~\cite{heckel_deep_2018}. The GRU architecture captures transient information between SAIs, while the deep decoder captures static information within SAIs. This two-stage compression pipeline exhibited competitive performance compared to JPEG-Pleno for light fields captured using Lytro camera. However, the introduction of the GRU module led to issues such as an unstable training procedure, memory overflow, and degraded performance when dealing with wide-baseline light fields.

Taking into account the limitations of the network designs in~\cite{heckel_deep_2018,jiang2022untrained}, we propose an implicit convolutional network backbone for light field representation, as depicted in Fig.~\ref{fig:workflow}. In this figure, the cuboids colored in blue, green, and orange represent convolutional kernels that serve distinct functionalities in each layer. Similar to~\cite{heckel_deep_2018}, the network's backbone consists of sequentially connected convolutional layers. During training, the network takes a volume of uniform noise $\epsilon$ as input, and it is supervised by a randomly selected SAI denoted as $I_{u,v}$. In this approach, the image is implicitly represented by the network's parameters, as follows:
\begin{equation}
    \Theta^* = \underset{\Theta} {\operatorname {arg\,min}} \, E(H_{\Theta}(\epsilon), I_{u,v}), \ \forall I_{u,v}\in L,
\label{eq:target_equ}
\end{equation}
where $H_{\Theta}(\cdot)$ represents the rendering procedure of the network, and $\Theta=\{K^{i},b^{i}\}$ are kernels weights and biases in each layer, $i$ is layer index. We use Mean Square Error (MSE) as loss function $E(\cdot)$ to supervise the training of the network.

When it comes to network design, we have opted for convolutional layers instead of fully connected layers for two main reasons: (1). The convolution operation is highly optimized for parallel computation and is more flexible for different input sizes. (2). Although fully connected layers can be used to construct Multi-Layer Perceptrons (MLPs) for light field representation, as demonstrated in works~\cite{mildenhall2020nerf,shi2023light}, the rendered output of MLPs is essentially individual pixels, making the output hard to be optimized in terms of losses such as Structural Similarity (SSIM)~\cite{wang2004image} or Learned Perceptual Image Patch Similarity (LPIP)~\cite{zhang2018unreasonable} that involve local regions of the rendered image. In contrary, convolutional layers produce feature maps or RGB images as outputs, enabling optimization with respect to metrics like SSIM or LPIP.
Regarding the network input, the reason of using uniform noise volume as input is twofold: (a). The noise does not contain any additional information, ensuring that it does not interfere with the learning of scene information during training. (b). The noise volume is generated using a pseudo random seed, when the encoder and decoder share the same seed, the transmission of the noise volume can be avoided. This helps conserve bandwidth and allows for efficient transmission in compression context.
Based on the choices of layer type and network input, the proposed backbone consists of six cascaded convolutional layers with kernel size $3\times3$, except for the last decoding layer, which converts channel number to 3 and has kernel size $1\times1$. To gradually increase the resolution of the feature maps, the first four layers are followed by bicubic upsampling operation (UP) with a scale factor of 2. Batch Normalization (BN) is added at the end of each layer to accelerate network convergence, except for the last layer. To expand the receptive field without increasing the number of network parameters, we set the kernel dilation rate to 2 for all intermediate layers. More details of the network backbone can be found in Tab.~\ref{table:architecture}.

\begin{table}[!thpb]
\begin{center}
\caption{The proposed network architecture. $k$, $s$, $d$ and \textit{in/out} represent the kernel size, the stride, kernel dilation size and the number of input/output channels, whereas `\textit{up2}', `\textit{BN}', `\textit{GELU}' and `\textit{Softmax}' represent bicubic upsampling by scale 2, Batch Normalization, activation functions GELU and Softmax. $c=c_{d}+c_{m}$ is the channel number of each layer, and 
$c_{d}, c_{m}$ are respectively channel numbers for descriptors and modulators.}
\begin{tabular}{c|ccccc}
\midrule[1.5pt]
\textbf{\textit{Layers}} & \textbf{\textit{k}} & \textbf{\textit{s}} & \textbf{\textit{d}} & \textbf{\textit{in/out}} & \textbf{\textit{input}}\\
\hline
\textit{$L_{1}$} & 3 & 1 & 1 & $c$/$c$ & noise $\epsilon$\\
\hline
\textit{\{up2,BN,GELU\}}& - & - & - & $c$/$c$ & \textit{$L_{1}$}\\
\hline
\textit{$L_{2}$} & 3 & 1 & 2 & $c$/$c$ & \textit{\{up2,BN,GELU\}}\\
\hline
\textit{\{up2,BN,GELU\}}& - & - & - & $c$/$c$ & \textit{$L_{2}$}\\
\hline
\textit{$L_{3}$}& 3 & 1 & 2 & $c$/$c$ & \textit{\{up2,BN,GELU\}}\\
\hline
\textit{\{up2,BN,GELU\}}& - & - & - & $c$/$c$ & \textit{$L_{3}$}\\
\hline
\textit{$L_{4}$}& 3 & 1 & 2 & $c$/$c$ & \textit{\{up2,BN,GELU\}}\\
\hline
\textit{\{up2,BN,GELU\}}& - & - & - & $c$/$c$ & \textit{$L_{4}$}\\
\hline
\textit{$L_{5}$}& 3 & 1 & 1 & $c$/$c$ & \textit{\{up2,BN,GELU\}}\\
\hline
\textit{\{BN,GELU\}}& - & - & - & $c$/$c$ & \textit{$L_{5}$}\\
\hline
\textit{Decoder}& 3 & 1 & 1 & $c$/3 & \textit{\{up2,BN,GELU\}}\\
\hline
\textit{Softmax}& - & - & - & 3/3 & \textit{Decoder}\\
\midrule[1.5pt]
\end{tabular}
\end{center}
\label{table:architecture}
\end{table}

\subsection{Complementary descriptor \& modulator design}
As previously introduced, the combination of GRU and deep decoder in~\cite{jiang2022untrained} enables a compact light field representation, but the utilization of GRU for modelling angular prior also leads to limitations such as memory overflow and degraded performance for wide-baseline data. We thus follow a different design philosophy for light field representation, which involves a single network and does not require additional modules.

Given a light field, it's obvious that all SAIs share similar scene content, and they also possess distinct visual elements such as occlusion and reflection patterns altered in terms of perspectives. 
When designing a network for compact representation, it should be able to store scene information for rendering and the rendering of each SAI should be controlled by the queried perspectives. 
To fullfil this requirement, we therefore defined two types of kernels in the network: \textbf{\textit{Descriptors}}, which store the scene description information and constitute the majority of the network's parameters, are repeatedly used when rendering every SAI. \textbf{\textit{Modulators}}, the auxiliary view-wise kernels indexed by angular coordinates $(u,v)$, will modulate the rendering process and are switched from one set to another when rendering different SAIs.
Like illustrated in Fig.~\ref{fig:workflow}, from the first to the second last layer of the network backbone, each layer $\{K^{i}, b^{i}\}$ is composed of descriptors (colored in blue) $K^{d_{i}}$ and modulators (colored in green and orange) $\{K^{m_{i}}_{u,v},b^{m_{i}}_{u,v}\}$: 
\begin{gather}
    K^{i} = K^{d_i}\oplus K^{m_i}_{u,v}, \ \
    b^{i} = b^{m_i}_{u,v},
\end{gather}
with $\oplus$ being concatenation operation in the last dimension. And $K^{d_i}$ and $K^{m_i}_{u,v}$ are tensors of sizes $k \times k \times C^{i}_{in} \times
C^{d_i}_{out}$ and $k \times k \times C^{i}_{in} \times C^{m_i,uv}_{out}$, where $k$ is the kernel size, and $C_{in}$ and $C_{out}$ are respectively the numbers of input and output channels. Thanks to this complementary kernel design, the network gets rid of an additional module for explicit angular prior modelling, making the overall architecture concise and effective.
Another advantage of using complementary kernel design is the computational resource reduction. The switchable kernel design makes the network generate one SAI in each forward pass, hence the memory consumption stays always at a low level.

The training of such a network involves the construction of a random SAI sampling stream. Specifically, as depicted on the right side of Fig.~\ref{fig:workflow}, in each iteration, a random SAI is selected from $U\times V$ light field views. The selected SAI along with its corresponding angular coordinates, forms a triplet $(u,v,I_{u,v})$. Modulators $(K^{m_{i}}_{u,v},b^{m_{i}}_{u,v})$ indexed by $(u,v)$ are then integrated into the network to work in tandem with descriptors for rendering $\hat{I}_{u,v}$. And $I_{u,v}$ serves as the ground truth for minimizing the reconstruction error. In the subsequent iteration, a new SAI is fed into the network, and the current modulators are replaced by the next set of modulators. It's noteworthy that the functionalities of scene description and rendering modulation for descriptors and modulators are automatically acquired during the training procedure.

\subsection{Allocation of modulator along angular directions}
\label{sec:orth_ang_rep}
For INR-based methods, the compression efficiency is largely decided by the number of parameters of the network. Although our adoption of descriptors and modulators preliminarily reduces the number of parameter for compact light field representation. The network may still suffer from the risk of parameter explosion: assuming we employ an $l$-layer network to represent an light field, its total number of parameters $N$ can be estimated approximately as follows:
\begin{equation}
    N \approx lk^{2}C_{in}(UVC^{m}_{out} + C^{d}_{out}),
\label{eq:total_number_1}
\end{equation}
where the number of parameter for modulators is proportional to $UV$ if we allocate a set of modulators to each SAI. In the case that the target light field has high angular resolution, the number of parameter for modulator will increase significantly and make the compression fail consequently.

To avoid parameter explosion for high angular resolution light fields meanwhile preserving good representation capability, instead of allocating modulators $\{K^{m_i}_{u,v},b^{m_i}_{u,v}\}$ to each angular position pair $(u,v)$, we propose to allocate modulators along two angular directions $u$ and $v$ by splitting them into two subsets $\{K^{m_i}_{u},b^{m_i}_{u}\}$ and $\{K^{m_i}_{v},b^{m_i}_{v}\}$ as follows:
\begin{gather}
    K^{m_i}_{u,v} = K^{m_i}_{u}\oplus K^{m_i}_{v}\\
    b^{m_i}_{u,v} = b^{m_i}_{u} + b^{m_i}_{v},
\end{gather}
where the channel number of $K^{m_i}_u$ and $K^{m_i}_v$ is half of that of $K^{m_i}_{u,v}$. Two subsets $\{K^{m_i}_{u},b^{m_i}_{u}\}$ and $\{K^{m_i}_{v},b^{m_i}_{v}\}$ are respectively represented by cuboids colored in green and orange in Fig.~\ref{fig:workflow}.
Such allocation along orthogonal directions is based on the observation that views in the same row exhibit similar variation mode in the horizontal direction, while those in the same column exhibit similar variation mode in the vertical direction.
Based on this allocation, the total number of parameter will be:
\begin{equation}
    N \approx lk^{2}C_{in}[\frac{1}{2}(U+V)C^{m}_{out} + C^{d}_{out}],
\label{eq:total_number_2}
\end{equation}
which means the number of parameter for modulators will be proportional to $\frac{1}{2}(U+V)$ instead of $UV$, implying a significant reduction of parameter particularly when dealing with high angular resolution light fields. Further discussion on the effectiveness of this allocation is given in Sec.~\ref{sec:kernel_design}.

\subsection{Decomposition of network kernel tensor}
\label{sec:kernel_decom}
As mentioned earlier, the INR-based method establishes a connection between light field compression and network compression. We can also leverage network compression techniques to further enhance the network's compactness. Recall that kernel weights $\{K^{d_i}, K^{m_i}_{u}, K^{m_i}_{v}\}$ are all four-dimension tensors, and employing suitable network compression techniques can help reduce the total number of parameters. 
In a related work \cite{shi2023light}, the authors applied model compression techniques to light field compression by introducing a rank-constrained NeRF \cite{mildenhall2020nerf} followed by network distillation. However, these techniques result in a complex training schedule and are primarily designed for fully-connected layers, hence are less preferable to our architecture. 
Inspired by the work \cite{qiu2018dcfnet} where the authors propose to decompose convolutional kernel tensors into the product of Fourier-Bessel (FB) bases \cite{abramowitz1964handbook} and corresponding weighting volume for parameter reduction purpose, we took similar operations on both descriptors and modulators by decomposing them into the shared base volume $B$ (yellow cylinder in Fig.~\ref{fig:workflow}) and coefficient volumes $\{W^{d_i}, W^{m_i}_{u}, W^{m_i}_{v}\}$ (blue, green and orange cylinders in Fig.~\ref{fig:workflow}), let us take descriptors $K^{d_i}$ as an example: 
\begin{equation}
    K^{d_i} = B \otimes W^{d_i},
\end{equation}
where $K^{d_i}$ is of size $k \times k \times C^{i}_{in} \times C^{d_i}_{out}$, $B$ is of size $k \times k \times r$, and $W^{d_i}$ is the coefficient volume of size $r \times C^{i}_{in} \times C^{d_i}_{out}$. The symbol $\otimes$ denotes the matrix multiplication and $r$ is the number of bases in $B$. 
The authors of \cite{qiu2018dcfnet} have shown that the Fourier-Bessel (FB) bases \cite{abramowitz1964handbook} are effective bases for compressing a network for image classification and denoising tasks. We therefore initialize $B$ with FB bases for faster convergence, with base number $r=6$.  we then update the bases during training to make them more specific to the scene being learned. Please note that a higher compression ratio can be achieved by using a smaller $r$. By following the network backbone design with complementary kernels and by employing the techniques of modulator allocation and kernel tensor decomposition, a light field can be compactly represented through a set of network parameters:
\begin{equation}
\Theta^{*} = \{B, W^{d_i}, W^{m_i}_{u}, W^{m_i}_{v}, b^{m_i}_{u}, b^{m_i}_{v}\}.
\end{equation}

\subsection{Quantization-aware training}
\label{sec:taq}
Besides the number of parameters required for representing a light field, the number of bit assigned to each parameter is also a significant factor in deciding compression efficiency. Although half precision (16 bits) has commonly been used in training deep learning frameworks, such a fixed-point scalar quantization with uniformly distributed centroids is still sub-optimal for the compression task. Here, we applied non-uniform quantization to each layer of the network for further network size reduction. More precisely, given a pre-defined number of centroids $n$ for each layer (except for the last decoding layer), when working on a certain layer $l_{i}$, we perform k-means clustering on the parameters $\{W^{d_i},W^{m_i}_{u},W^{m_i}_{v},b^{m_i}_{u},b^{m_i}_{v}\}$ to obtain $n$ centroids $\gamma_{i}$, and these centroids will be updated to minimize reconstruction error:
\begin{equation}
    \gamma_{i}^* = \underset{\gamma_{i}} {\operatorname {arg\,min}} \, E(H_{\Theta}(\epsilon), I_{u,v}), \ \Theta_{i} \in \gamma_{i}, \forall I_{u,v}\in L 
\label{eq:target_equ2}
\end{equation}
As the quantization error will accumulate throughout the network if all layers are simultaneously quantized, we adopted similar solution as in~\cite{fan2020training,jiang2022untrained,shi2023light}, which proposes to quantize network parameters layer by layer, i.e. after quantizing the current layer, we fix the parameters of this layer with the learned codewords $\gamma_{i}^*$, and continue to finetune all consecutive layers. We perform 16-bit uniform quantization on the last decoding layer, as we found that non-uniformly quantize the last layer with a small $n$ brings significant quality degradation. The bases $B$ are likewise quantized using uniform 16-bit quantization for better precision. In additional to uniform quantization with learned centroids, we also perform lossless entropy coding (Huffman coding) for further model compression. The quantized parameters of network will be transmitted from the encoder to the decoder, along with corresponding codewords at a cost of $n\times 32$ bits, with each codeword being encoded using 32 bits.

\section{Experimental Settings}
\subsection{Training details}
The global schedule consists of two phases: the training phase and the quantization phase. Both phases utilize a learning rate of 0.01. The training phase involves 12 epochs, with each epoch defined as all SAIs being used 500 times. In each iteration, 5 SAIs are fed into the network to calculate the averaged loss. Let us note that for a network-based light field representation, more training iteration means better performance, people can hence use a smaller training epoch for saving time or a larger epoch number for better reconstruction quality. While in the quantization stage, we define 1 epoch as all SAIs being involved 200 times. After quantizing each layer, we perform fine-tuning on all consecutive layers for 1 epoch The whole framework is implemented in Pytorch deep learning framework and trained on a single GPU of type Nvidia Titan RTX having 24GB memory. Both encoding and decoding time will be analyzed in Sec.~\ref{sec:memory_encoding_decoding}.

\subsection{Test datasets}
We take four synthetic scenes \textit{`boxes', `sideboard', `cotton', `dino'} from the HCI dataset \cite{honauer2016dataset} and four real-world scenes \textit{`Bikes', `Danger', `FountainVincent2', `StonePillarsOutside'} from the EPFL light field dataset \cite{EPFLLFdataset} as test data. Both datasets are widely used by the light field research community and have distinct but representative characteristics. 

The four real-world light fields are captured with plenoptic camera Lytro Illum~\cite{ng2005light} with narrow baseline, they have spatial resolution $432\times 624$ and angular resolution $13\times 13$, due to the vignetting effect, we take the central $9\times 9$ SAIs in our test. The introduction of micro-lens array reduces the luminance arriving at the sensor, light fields captured by Lytro Illum are generally noisy, which can be used to validate the robustness against noise for the compared methods.
In contrary, the four synthetic scenes are rendered using the 3D graphics software blender \cite{blender}, they have spatial resolution $512\times 512$ and angular resolution $9\times 9$. The synthetic data mainly simulates the light fields captured by camera array, hence they have lower noise level and a wider baseline. This type of data is able to assess the performance on light fields having large disparity range for each method.

\subsection{Method configurations}
We evaluate the performance of our proposed method for the task of compression, and compare it with SOTA methods that represent the recent trends in this domain, including classic video coding standard HEVC-Lozenge \cite{HEVCstandard,rizkallah2016impact}, learning-based video compression schemes HLVC \cite{yang2020Learning} and RLVC \cite{yang2021learning}, solutions dedicated to light field compression task such as JPEG-Pleno \cite{jpegpleno}, and the most recent INR-based methods DDLF \cite{jiang2022untrained} and QDLR-NeRF \cite{shi2023light}. We use official codes for all compared methods in our experiments, and each one is configured as follows:
\begin{itemize}
\item We use HEVC in version HM-16.10 in our test. Concerning the configuration of GOP and base QPs, we adopt a GOP of 4 as in~\cite{jiang2022untrained,shi2023light} and set QP = $\{20,22,24,28,32,36\}$ for real-world light fields and QP = $\{18,22,26,30,34\}$ for synthetic ones.
\item For two learning-based video compression schemes HLVC and RLVC, both of them have an optional hyper-parameter $\lambda=\{256,512,1024,2048\}$ to control the trade-off between bitrate and distortion. The method HLVC adopts a default GOP of 10 to realize frame prediction via three hierarchical quality layers, while for RLVC, 6 P-frames are bidirectionally encoded with a GOP of 13.
\item The software version of JPEG Pleno we use is the Verification Model 2.0 in the WaSP mode, and we use disparity maps predicted by~\cite{shi2019depth} in the compression process.
\item For the method QDLR-NeRF, as both tensor rank $r$ and number of centroid $n$ for quantizatin can control the size of model, we use four different ranks $r=\{40,70,90,150\}$ with a fixed number of centroids $n=256$ to have medium bitrate, then reduce the number of centroid to $n=\{128,64,32\}$ with a fixed rank $r=40$ for low bitrate.
\item The method DDLF is with parameters $(z_a,z_s)=\{(15,30),(20,40),(25,50),(30,60)\}$ and 256 centroids in its architecture, where $(z_a,z_s)$ denote respectively the channel number of the input spatial and angular code vectors, both handcrafted and neural-based upsamplings (i.e.~pixel shuffle) are involved for having a wide range of bitrate.
\item Finally, for our method, we alter the channel number for modulator $c_{m}$ and for descriptor $c_{d}$ in each layer for different bitrates. More precisely, we use $(c_m,c_{d})=\{(2,48),(2,63),(2,78),(2,93),(2,123),(2,153),(2,183)\}$ in our network. Though a small rank $r$ and centroid number $n$ can decrease the bitrate, they will severely degrade the compression quality, hence we use $r=6$ and $n=256$ in our test. 
\end{itemize}

\begin{figure*}[htbp]
\centering 
\subfigure{\label{fig:rd_bikes}\includegraphics[width=0.4\linewidth]{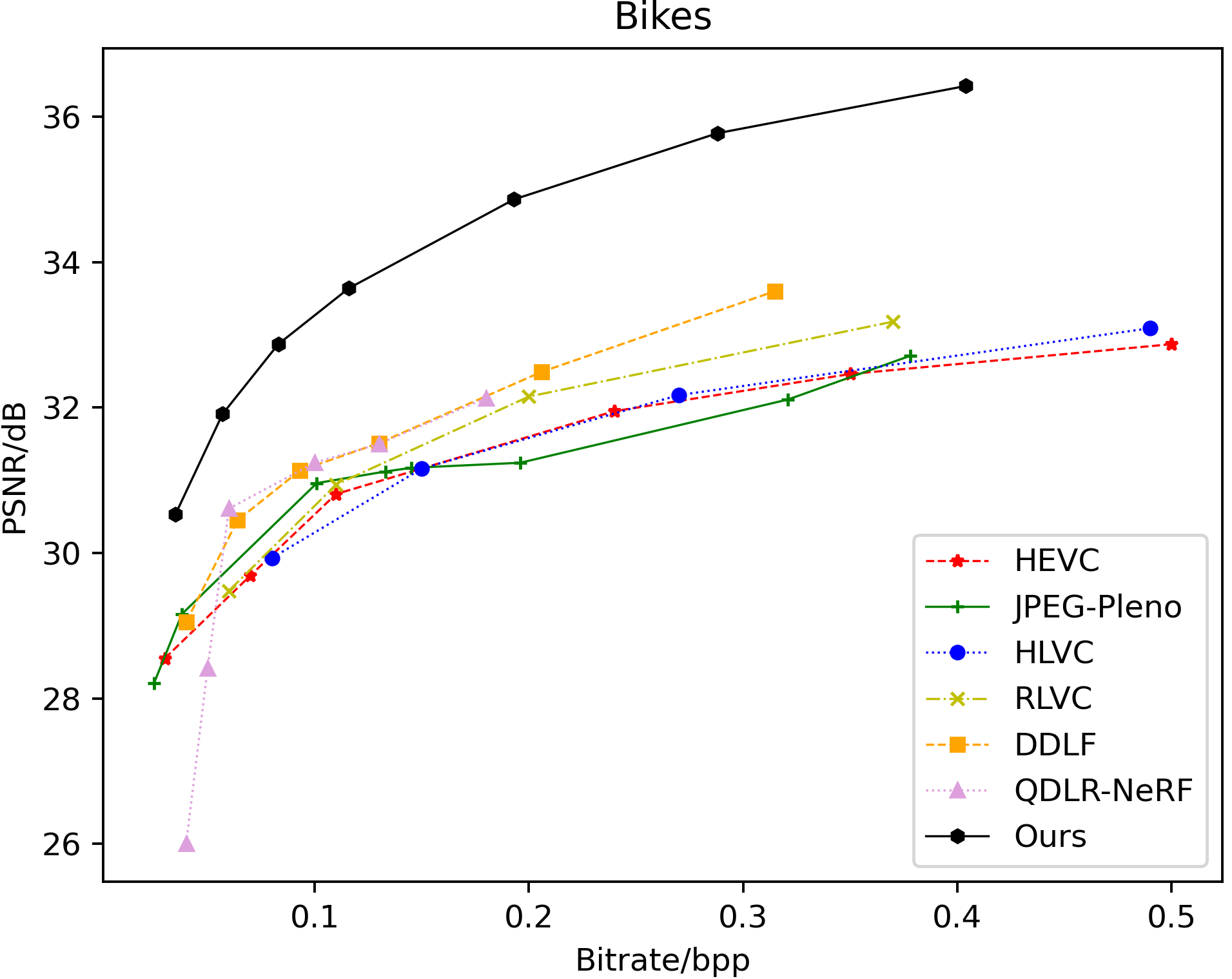}}
\hspace{+0.6cm}
\subfigure{\label{fig:rd_danger}\includegraphics[width=0.4\linewidth]{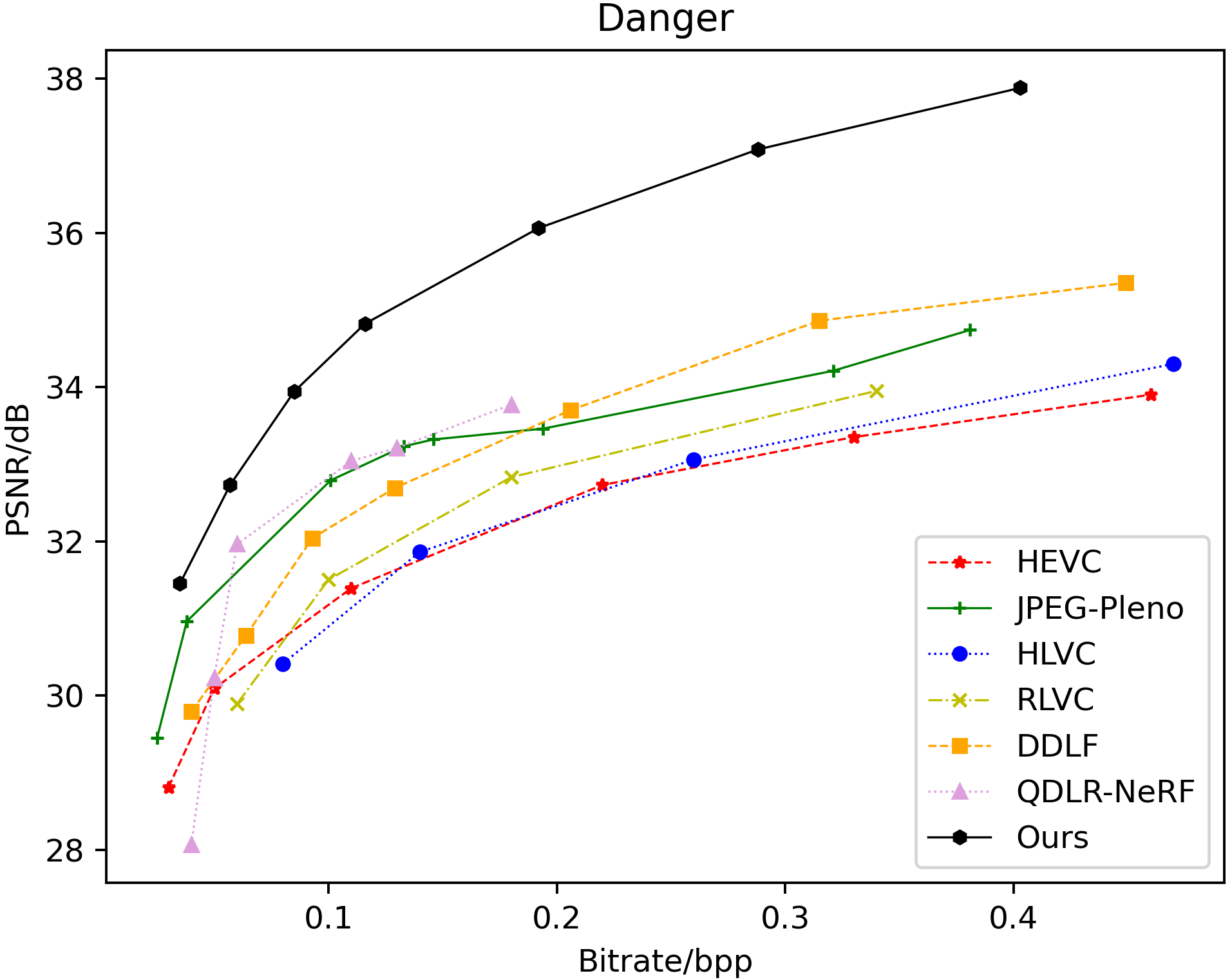}}
\\\vspace{-0.2cm}
\subfigure{\label{fig:rd_fountain}\includegraphics[width=0.4\linewidth]{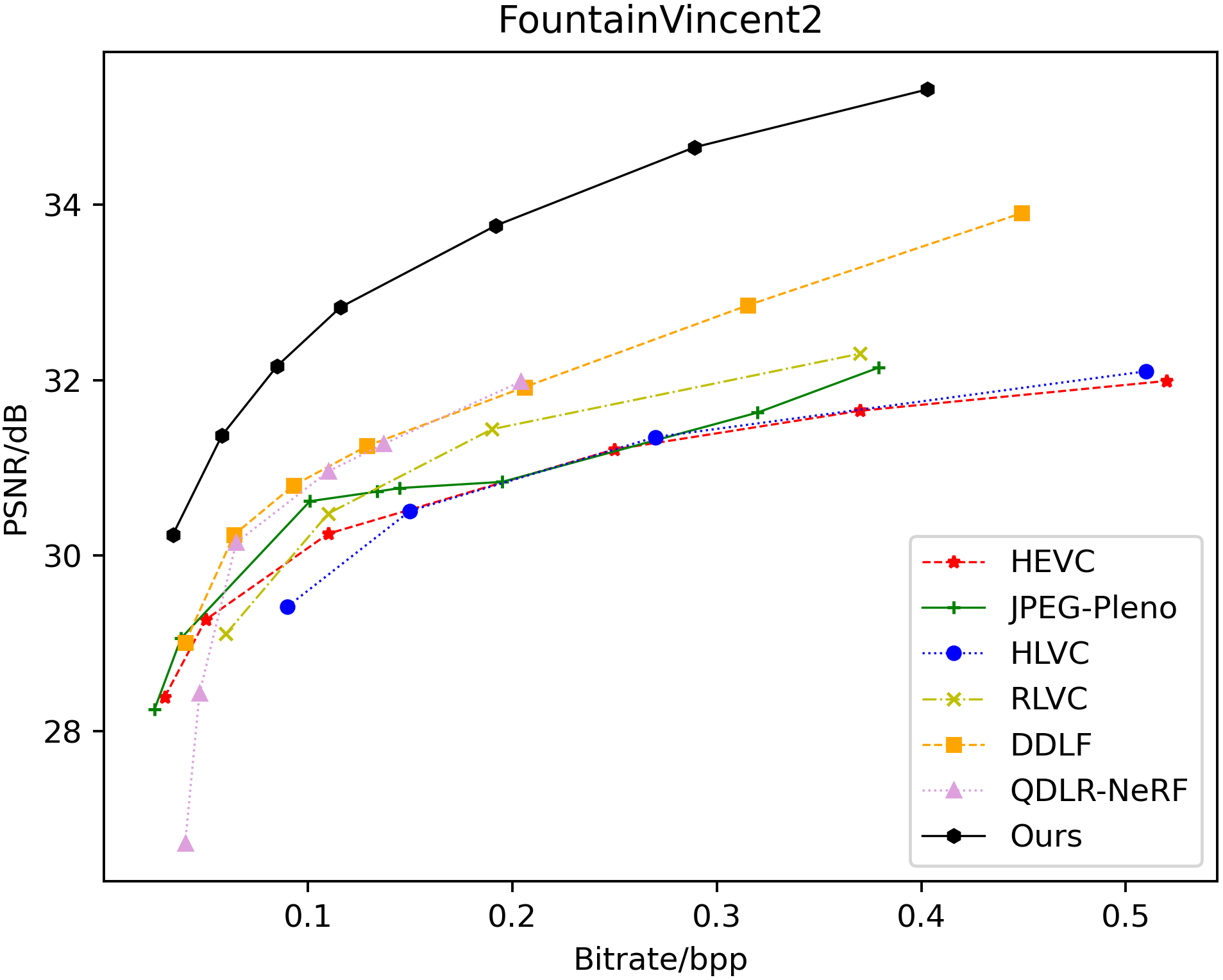}}
\hspace{+0.6cm}
\subfigure{\label{fig:rd_stone}\includegraphics[width=0.4\linewidth]{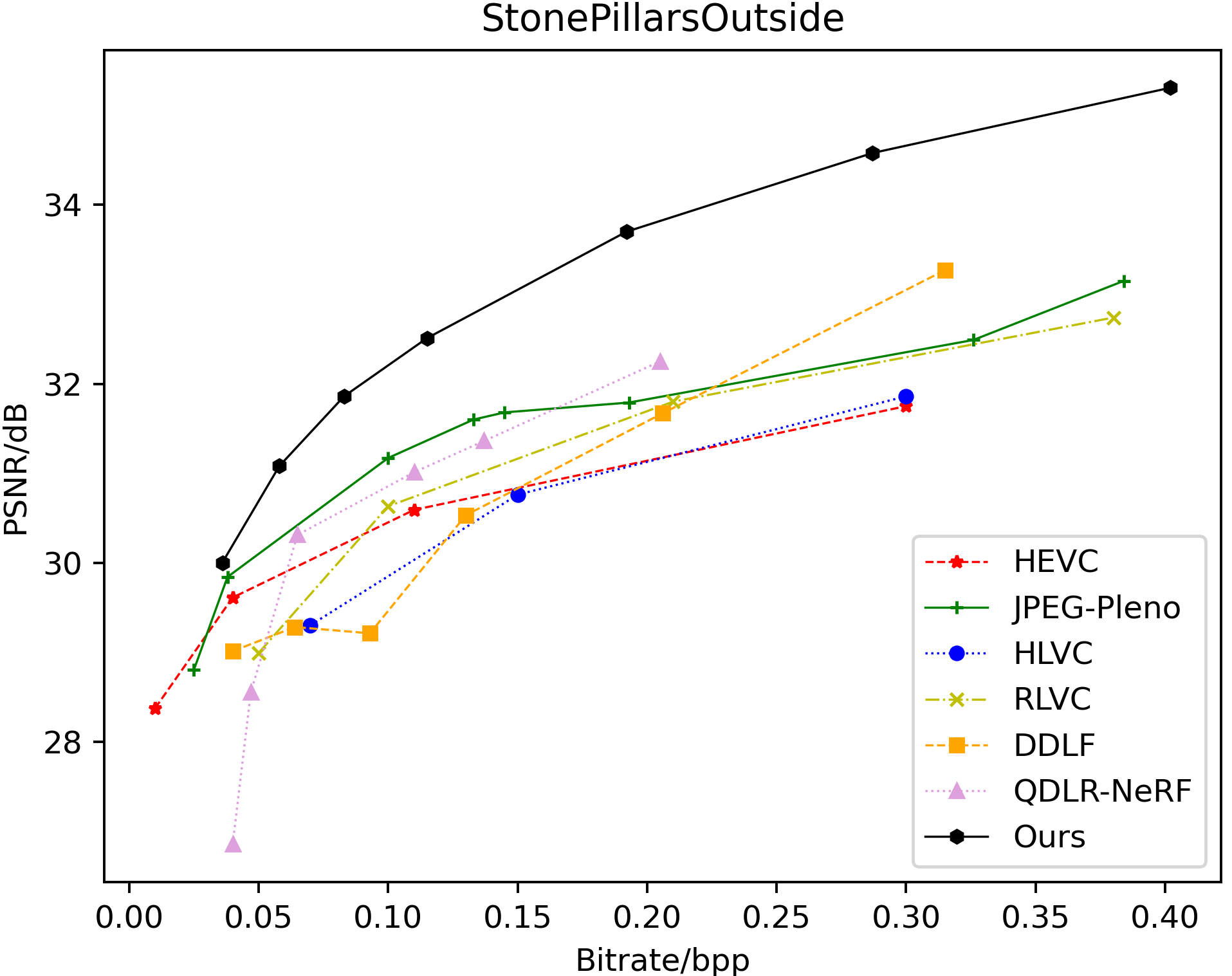}}
\\\vspace{-0.2cm}
\subfigure{\label{fig:rd_boxes}\includegraphics[width=0.4\linewidth]{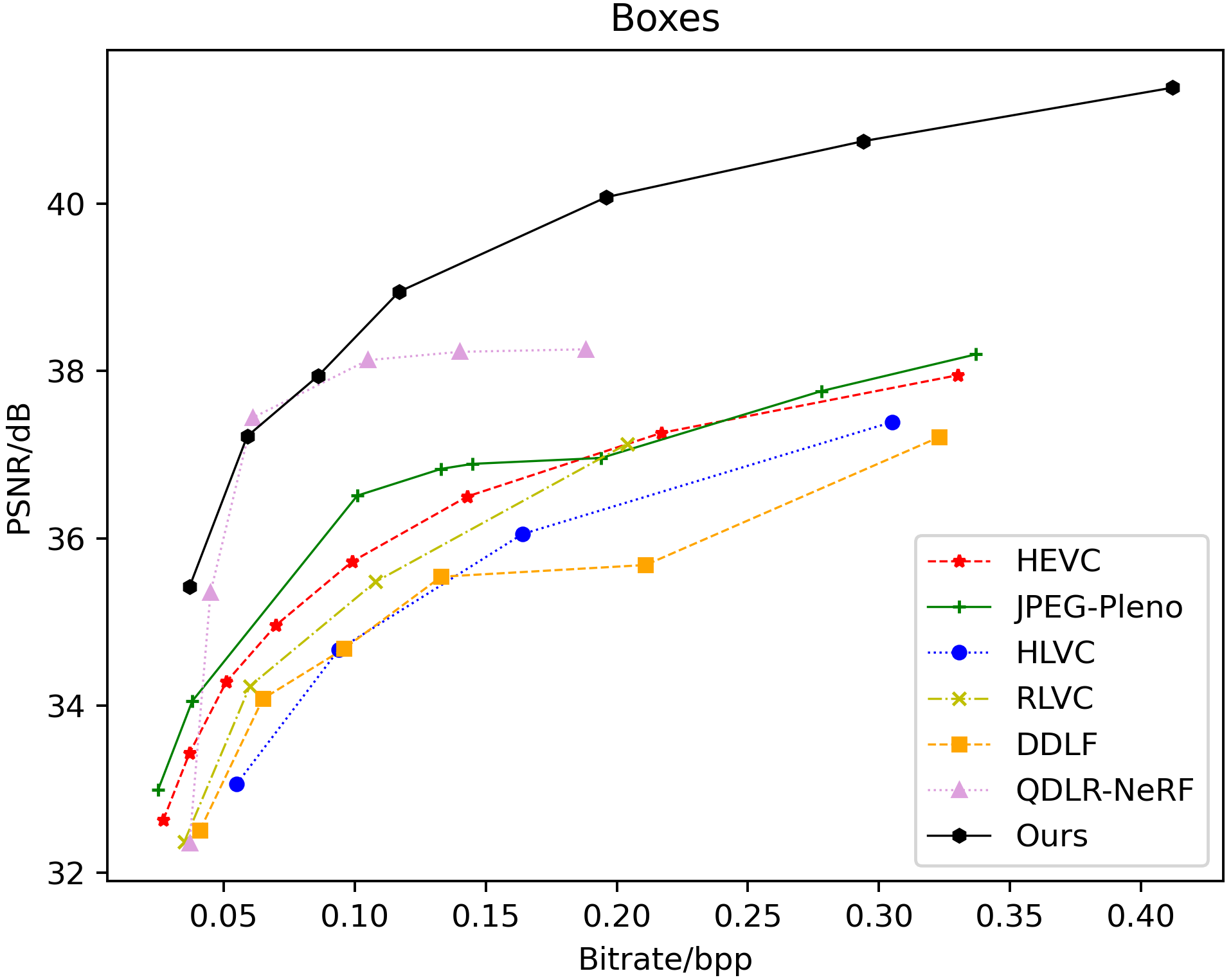}}
\hspace{+0.6cm}
\subfigure{\label{fig:rd_sideboard}\includegraphics[width=0.4\linewidth]{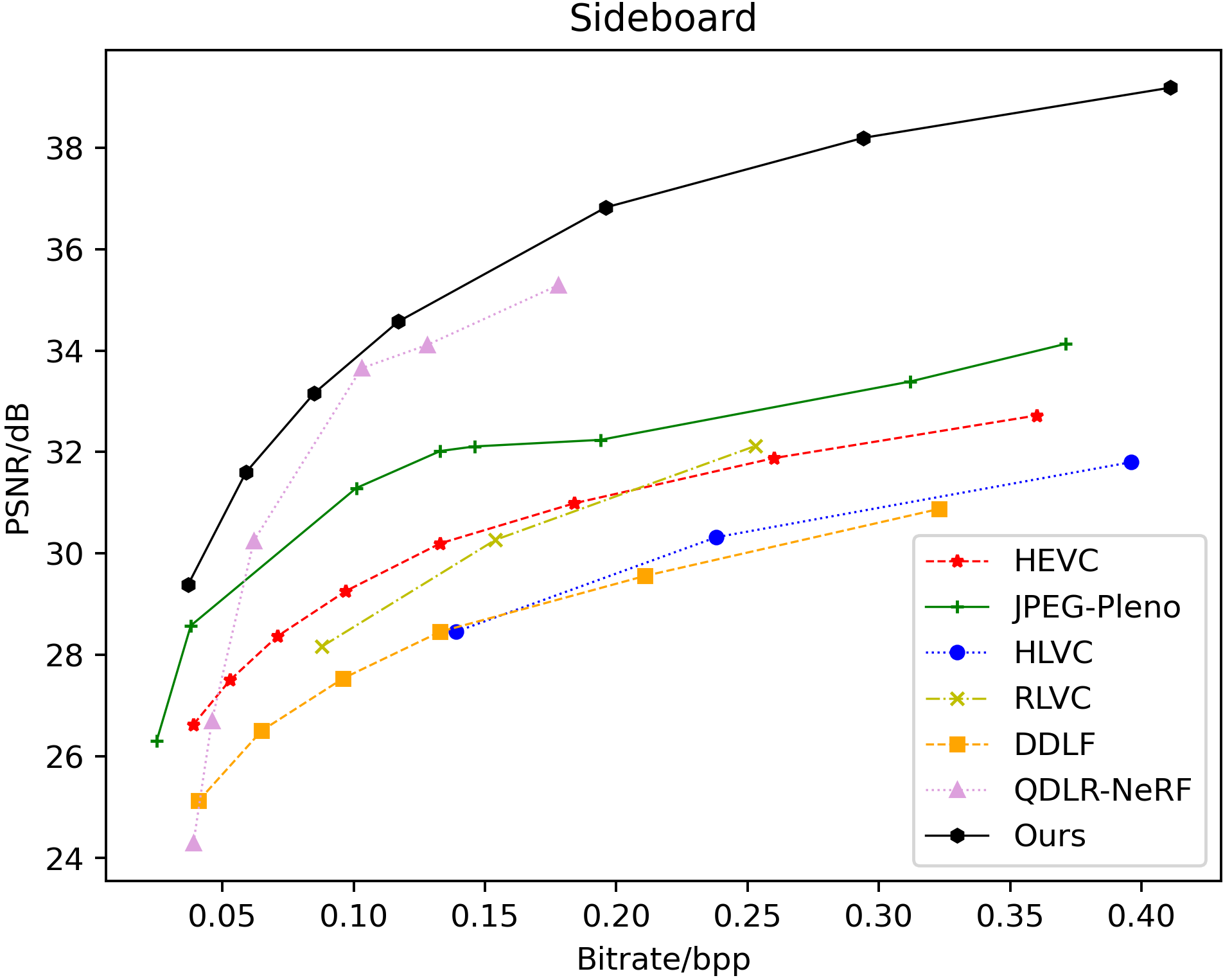}}
\\\vspace{-0.2cm}
\subfigure{\label{fig:rd_dino}\includegraphics[width=0.4\linewidth]{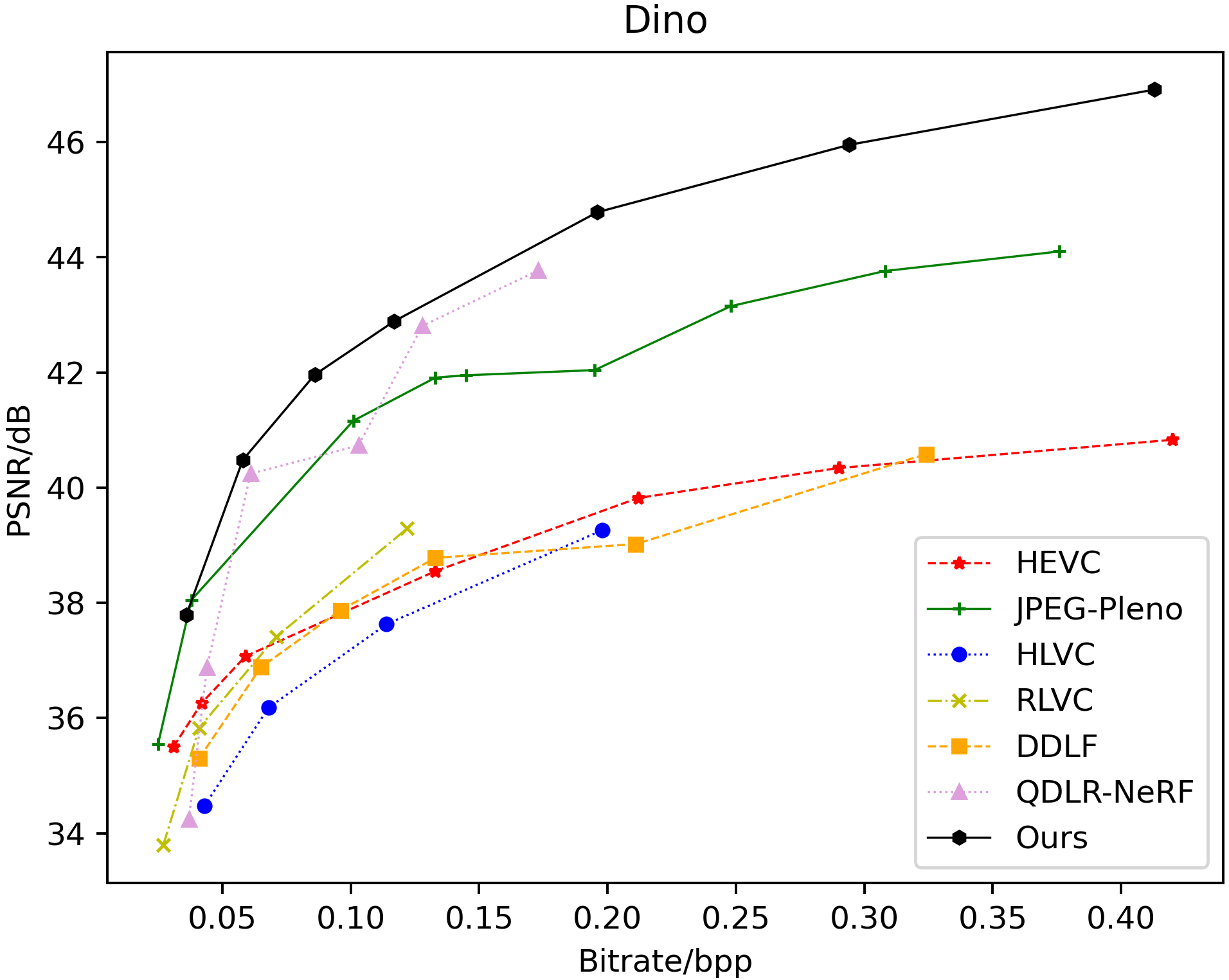}}
\hspace{+0.6cm}
\subfigure{\label{fig:rd_cotton}\includegraphics[width=0.4\linewidth]{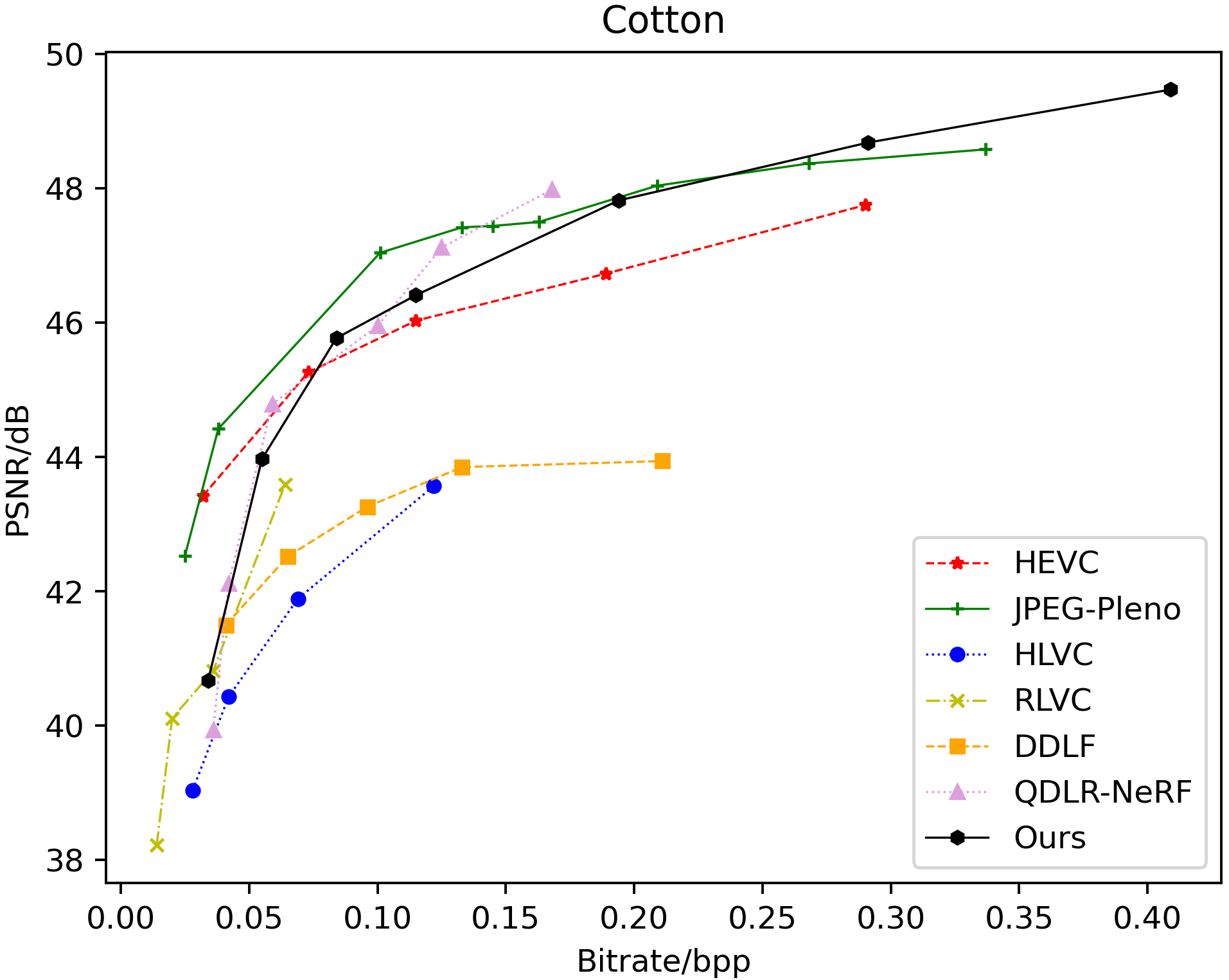}}
\vspace{-0.2cm}
\caption{Rate-distortion curves of \textit{'Bikes'}, \textit{'Danger'}, \textit{'FountainVincent2'}, \textit{'StonePillarsOutside'}, \textit{'Boxes'}, \textit{'Sideboard'}, \textit{'Dino'} and \textit{'Cotton'} for HEVC-Lozenge \cite{HEVCstandard,rizkallah2016impact}, JPEG-Pleno \cite{jpegpleno}, HLVC \cite{yang2020Learning}, RLVC \cite{yang2021learning}, DDLF \cite{jiang2022untrained}, QDLR-NeRF \cite{shi2023light} and ours.
}
\label{fig:rate_dist}
\end{figure*}

\section{Experimental Results}

\begin{table}
\centering
\caption{BD-PSNR gains with respect to HEVC baseline. The best results are in bold, and the second best results are underlined. *All involved methods cover a bpp range 0-0.45, except for the method QDLR-NeRF, which covers a bpp range 0-0.2.}
\label{table:DB_PSNR}
\tabcolsep=0.05cm
% \resizebox{0.68\textwidth}{!}{
\begin{tabular}{c|c c c c c c c c |c}
\hline
\hline
% \parbox[t]{3mm}{\multirow{7}{*}{\rotatebox[origin=c]{90}{\footnotesize BD-PSNR Gain}}} &
\footnotesize Methods  & \footnotesize boxes & \footnotesize sideboard & \footnotesize cotton  & \footnotesize dino & \footnotesize bikes & \footnotesize danger & \footnotesize SPO & \footnotesize FV2 & \footnotesize Average\\
% \hhline{-----}
\hhline{----------}
\footnotesize JPEG-P\cite{jpegpleno} & \footnotesize 0.46  & \footnotesize 1.77 & \footnotesize \textbf{0.99} & \footnotesize \underline{2.74} & \footnotesize 0.20  &  \footnotesize 1.38 & \footnotesize \underline{0.58} & \footnotesize 0.30 & \footnotesize 1.05\\
% \hhline{------}
\footnotesize HLVC\cite{yang2020Learning} & \footnotesize -0.78  & \footnotesize -1.37 & \footnotesize -3.29 & \footnotesize -1.06 & \footnotesize 0.01  &  \footnotesize 0.01 & \footnotesize -0.15 & \footnotesize -0.07 & \footnotesize -0.84\\
% \hhline{------}
\footnotesize RLVC\cite{yang2021learning} & \footnotesize -0.47  & \footnotesize -0.02 & \footnotesize -2.71 & \footnotesize -0.15  & \footnotesize 0.35  &  \footnotesize 0.29 & \footnotesize 0.20 & \footnotesize 0.32 & \footnotesize -0.27\\ 
% \hhline{------}
\footnotesize DDLF\cite{jiang2022untrained} & \footnotesize -1.00  & \footnotesize -1.75 & \footnotesize -3.14 & \footnotesize -0.31 & \footnotesize \underline{0.69}  & \footnotesize  0.93 & \footnotesize -0.01 & \footnotesize \underline{0.90} & \footnotesize -0.46\\
% \hhline{-------}
\footnotesize Q-NeRF*\cite{shi2023light}  & \footnotesize \underline{2.00}  & \footnotesize \underline{2.72} & \footnotesize -0.11 & \footnotesize 2.58 & \footnotesize 0.36  &  \footnotesize \underline{1.53} & \footnotesize 0.11 & \footnotesize 0.38 & \footnotesize \underline{1.20}\\
\hhline{----------}
 % \cmidrule{2-11}
\footnotesize \textbf{Ours} &\footnotesize  \textbf{2.74}  & \footnotesize \textbf{4.84} & \footnotesize \underline{0.21} &\footnotesize \textbf{4.34} & \footnotesize \textbf{1.88}  &\footnotesize \textbf{3.20} & \footnotesize \textbf{1.93} &\footnotesize \textbf{2.50} & \footnotesize \textbf{2.71}\\
\hline
\hline
\end{tabular}
% }
\end{table}

\begin{figure*}[!ht]
    \centering
    \setlength{
    \tabcolsep}{1pt}
    \centering
\resizebox{1.\textwidth}{!}{
\begin{tabular}{ccccccc}
\textbf{\textit{GT view ($I_{5,5}$)}} & HEVC\cite{HEVCstandard,rizkallah2016impact}& JPEG-Pleno\cite{jpegpleno} & RLVC\cite{yang2021learning} & DDLF\cite{jiang2022untrained} & QDLR-NeRF\cite{shi2023light} & Ours\\
\includegraphics[width=0.16\linewidth]{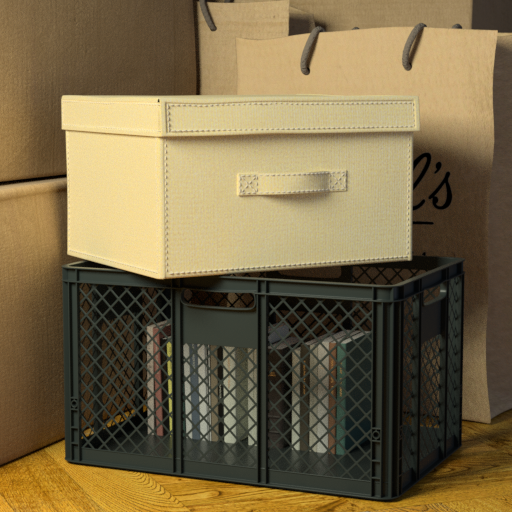} & 
\includegraphics[width=0.16\linewidth]{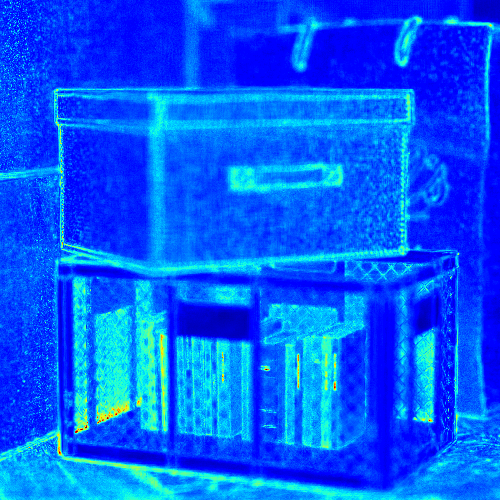} &
\includegraphics[width=0.16\linewidth]{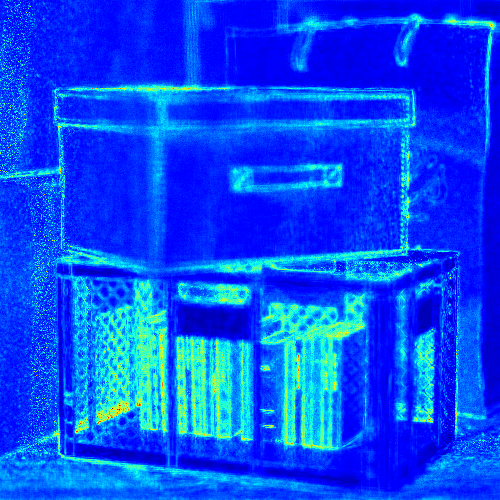} &
\includegraphics[width=0.16\linewidth]{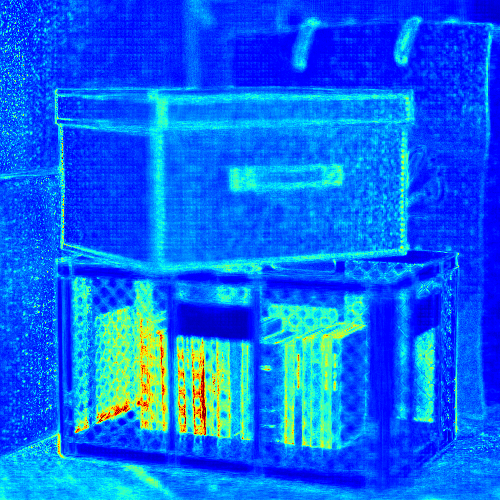} &
\includegraphics[width=0.16\linewidth]{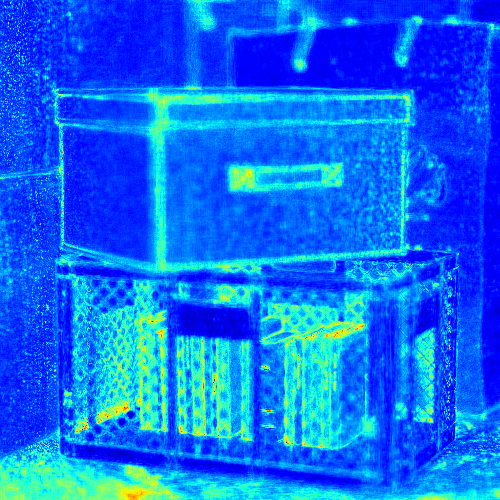} &
\includegraphics[width=0.16\linewidth]{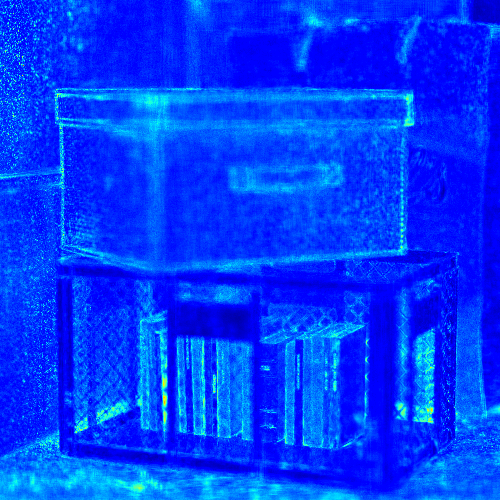} &
\includegraphics[width=0.16\linewidth]{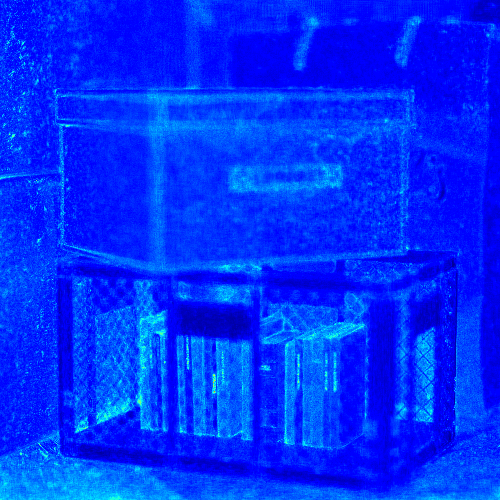}\\
\textbf{\textit{boxes}} & 36.51dB-0.14bpp & 36.82dB-0.13bpp & 35.53dB-0.11bpp & 35.48dB-0.13bpp & 38.13dB-0.11bpp & 38.84dB-0.12bpp\\
\includegraphics[width=0.16\linewidth]{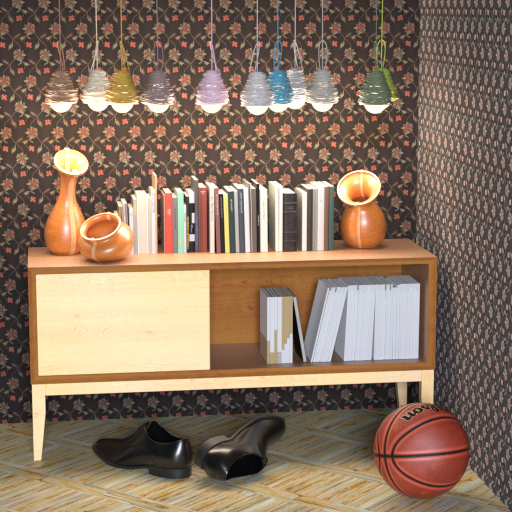} & 
\includegraphics[width=0.16\linewidth]{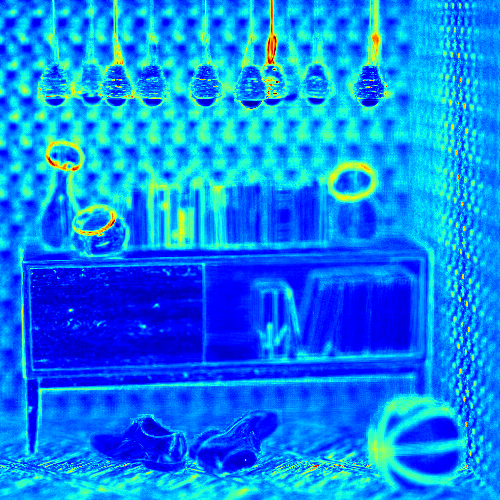} &
\includegraphics[width=0.16\linewidth]{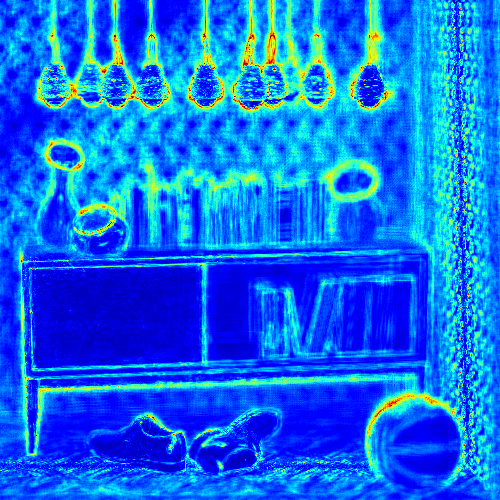} &
\includegraphics[width=0.16\linewidth]{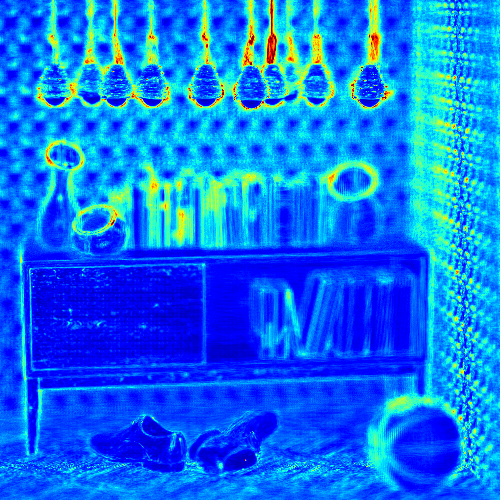} &
\includegraphics[width=0.16\linewidth]{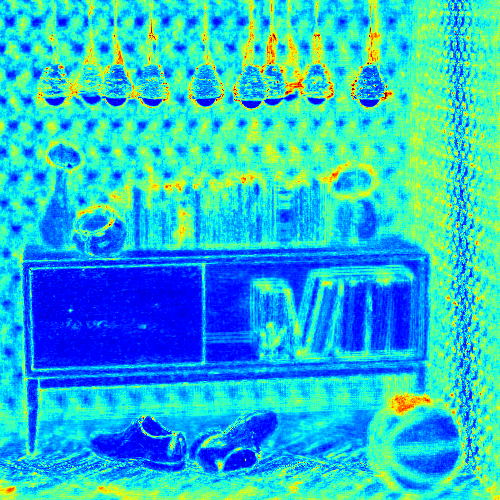} &
\includegraphics[width=0.16\linewidth]{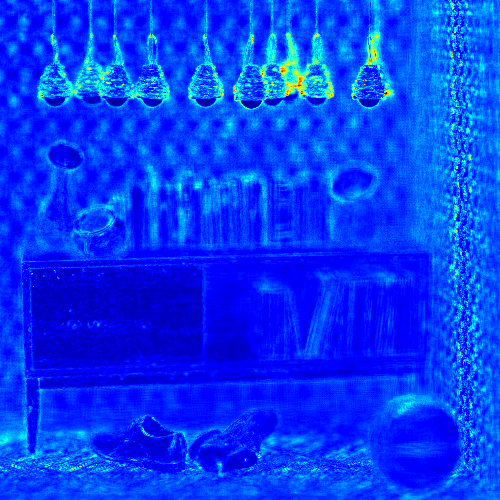} &
\includegraphics[width=0.16\linewidth]{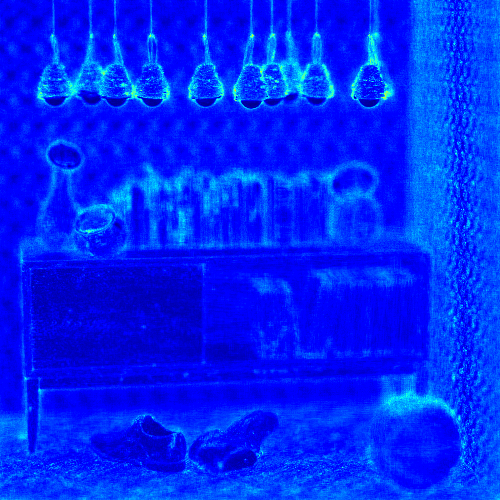}\\
\textbf{\textit{sideboard}} & 31.88dB-0.26bpp & 32.20dB-0.19bpp & 32.15dB-0.25bpp & 29.44dB-0.21bpp & 35.29dB-0.18bpp & 36.67dB-0.20bpp\\
\includegraphics[width=0.16\linewidth]{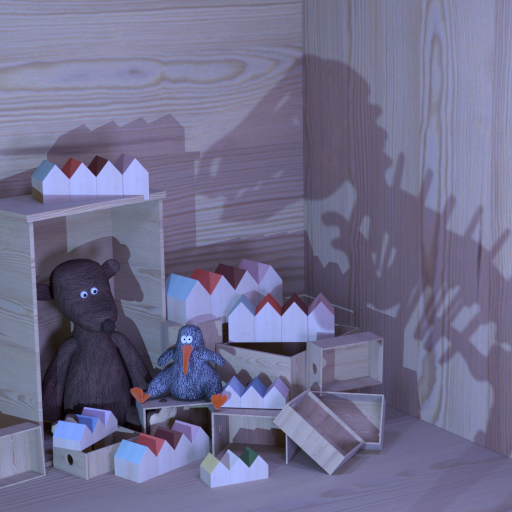} & 
\includegraphics[width=0.16\linewidth]{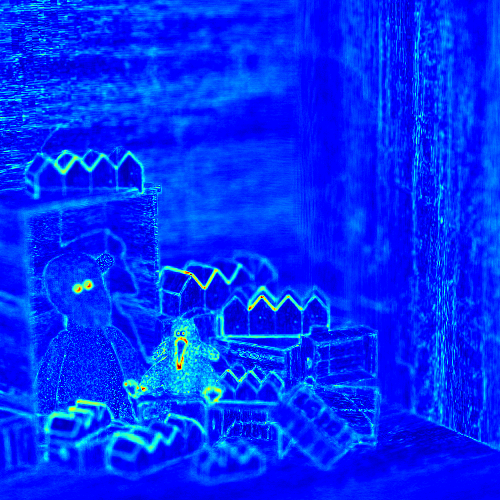} &
\includegraphics[width=0.16\linewidth]{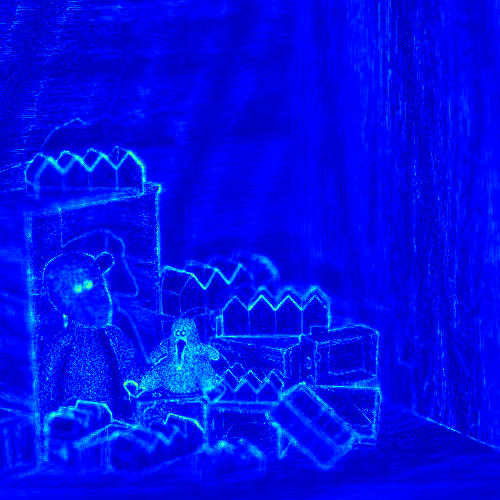} &
\includegraphics[width=0.16\linewidth]{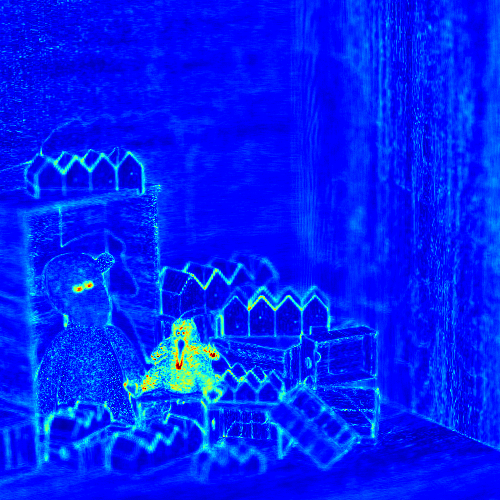} &
\includegraphics[width=0.16\linewidth]{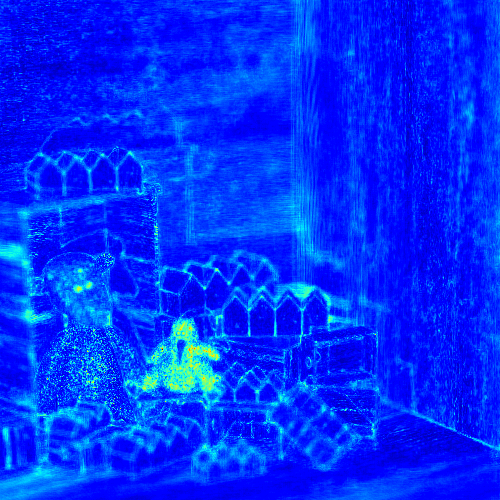} &
\includegraphics[width=0.16\linewidth]{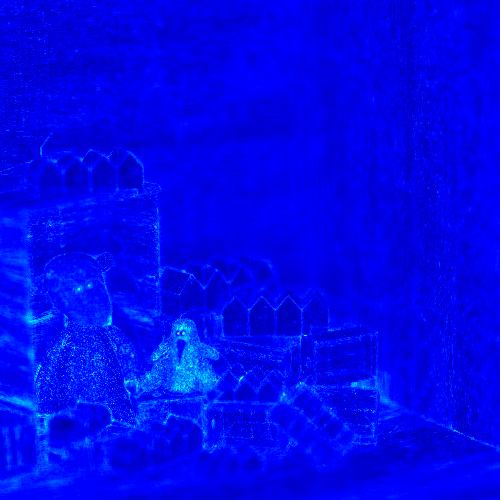} &
\includegraphics[width=0.16\linewidth]{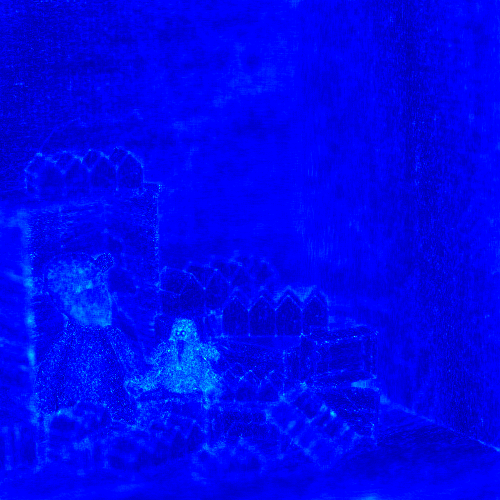}\\
\textbf{\textit{dino}} & 38.54dB-0.11bpp & 41.91dB-0.13bpp &  39.29dB-0.12bpp &  38.78dB-0.13bpp & 42.83dB-0.13bpp & 42.89dB-0.12bpp\\
\includegraphics[width=0.16\linewidth]{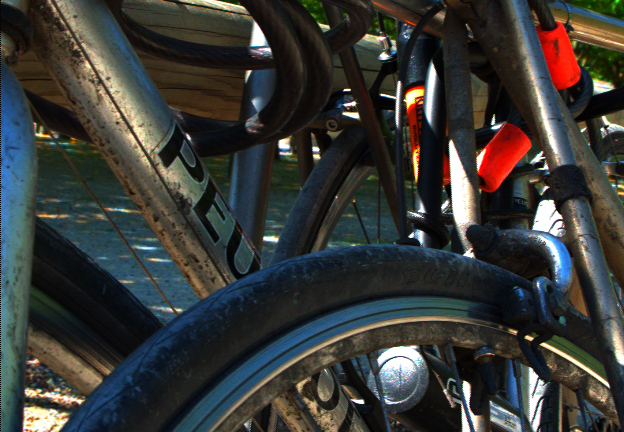} & 
\includegraphics[width=0.16\linewidth]{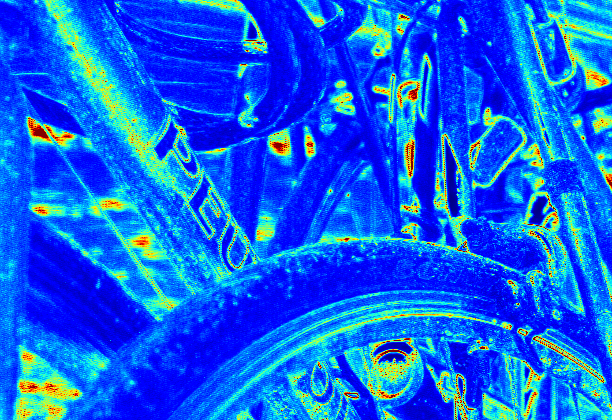} &
\includegraphics[width=0.16\linewidth]{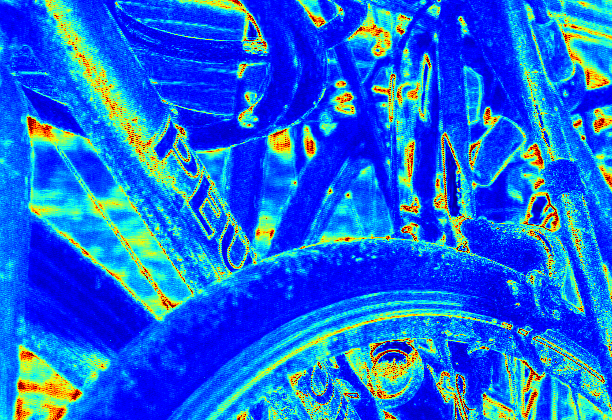} &
\includegraphics[width=0.16\linewidth]{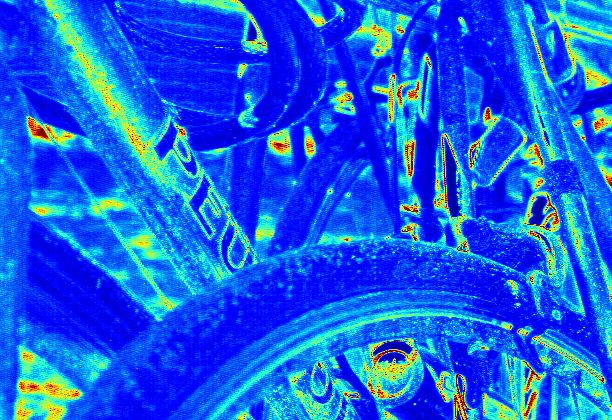} &
\includegraphics[width=0.16\linewidth]{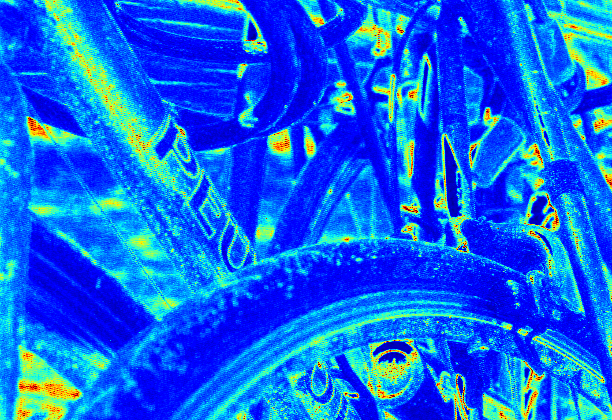} &
\includegraphics[width=0.16\linewidth]{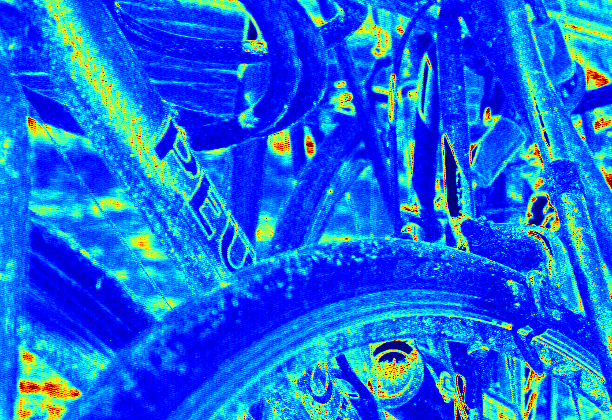} &
\includegraphics[width=0.16\linewidth]{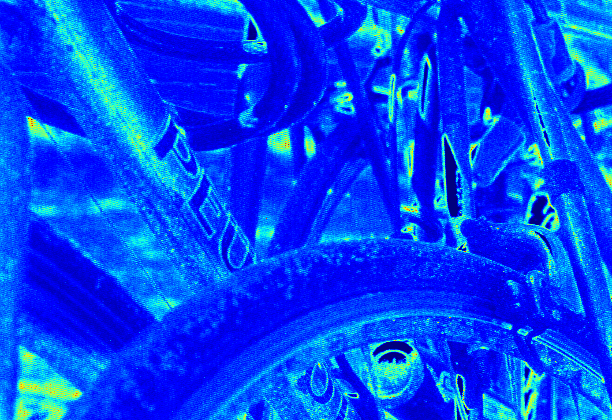}\\
\textbf{\textit{Bikes}} & 31.95dB-0.24bpp & 30.94dB-0.20bpp & 31.77dB-0.20bpp & 32.36dB-0.21bpp & 32.02dB-0.18bpp & 34.74dB-0.19bpp\\
\includegraphics[width=0.16\linewidth]{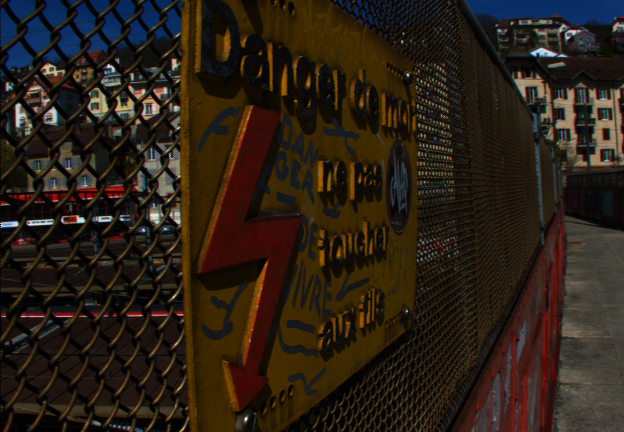} &
\includegraphics[width=0.16\linewidth]{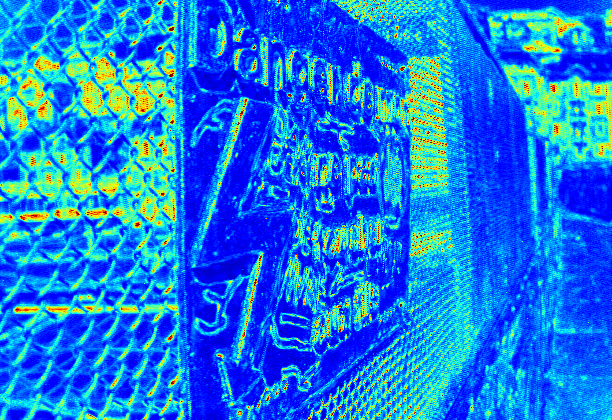} &
\includegraphics[width=0.16\linewidth]{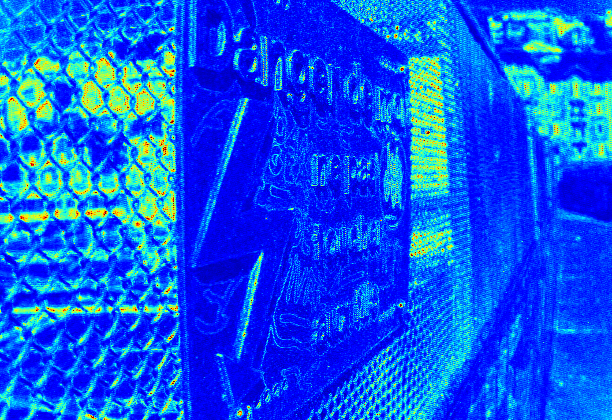} &
\includegraphics[width=0.16\linewidth]{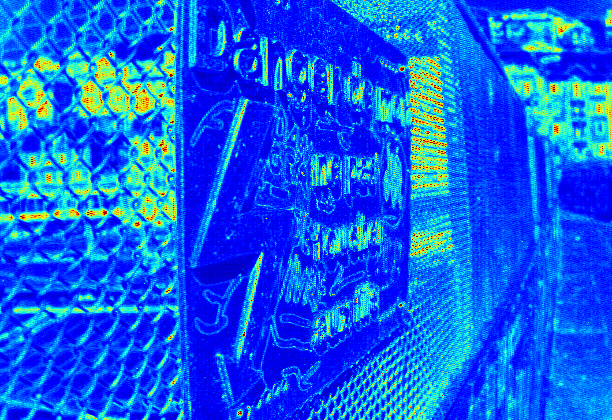} &
\includegraphics[width=0.16\linewidth]{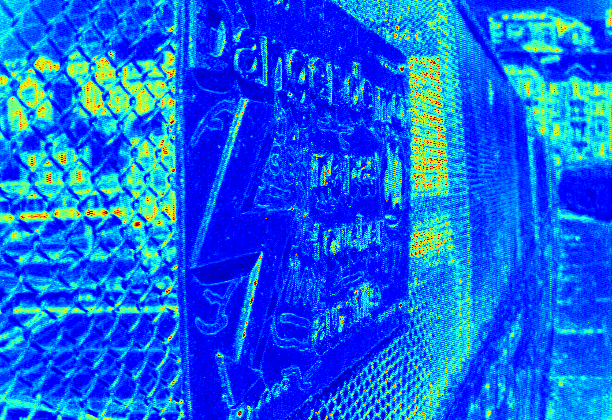} &
\includegraphics[width=0.16\linewidth]{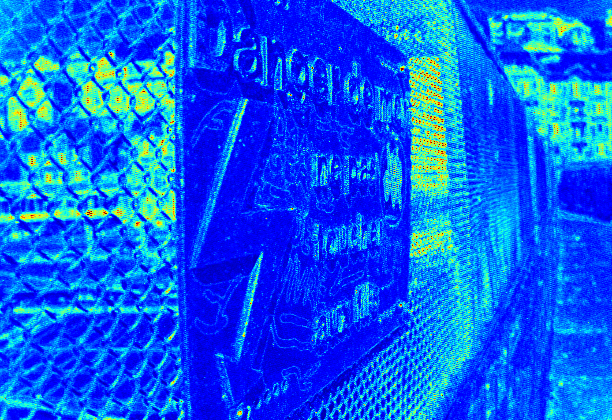} &
\includegraphics[width=0.16\linewidth]{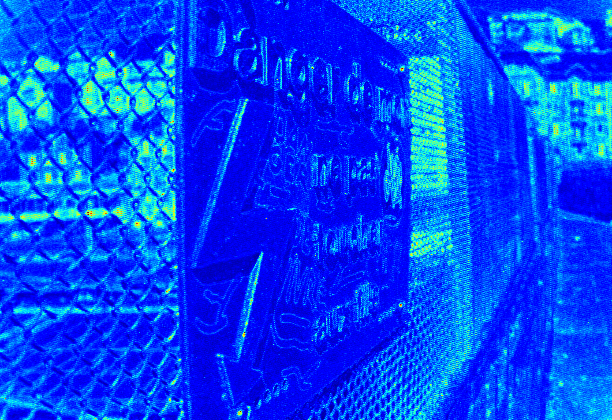}\\
\textbf{\textit{Danger}} & 31.39dB-0.11bpp & 33.19dB-0.13bpp & 32.78dB-0.18bpp & 32.62dB-0.13bpp & 33.20-0.13bpp & 34.77dB-0.12bpp\\
\includegraphics[width=0.16\linewidth]{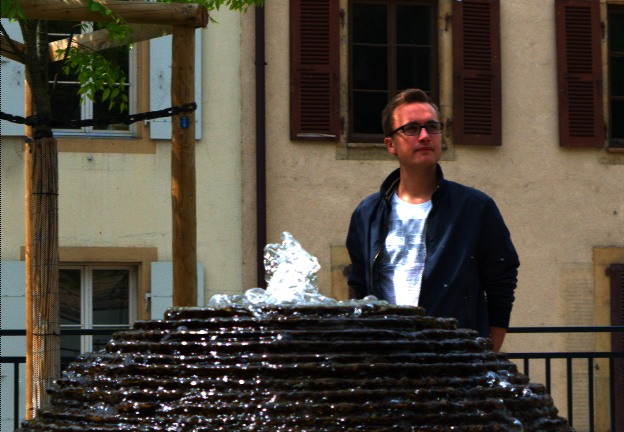} & 
\includegraphics[width=0.16\linewidth]{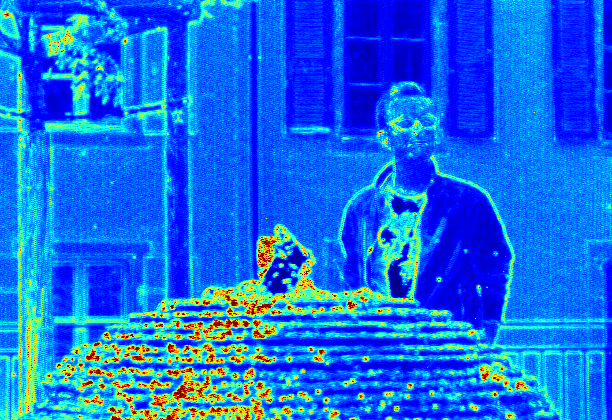} &
\includegraphics[width=0.16\linewidth]{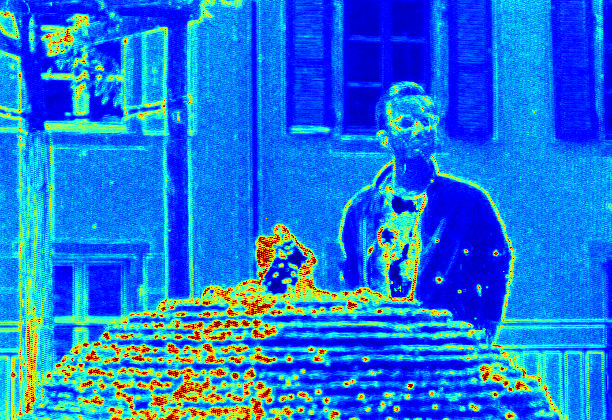} &
\includegraphics[width=0.16\linewidth]{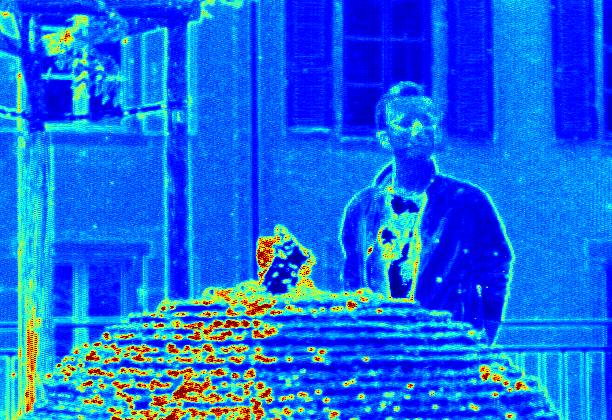} &
\includegraphics[width=0.16\linewidth]{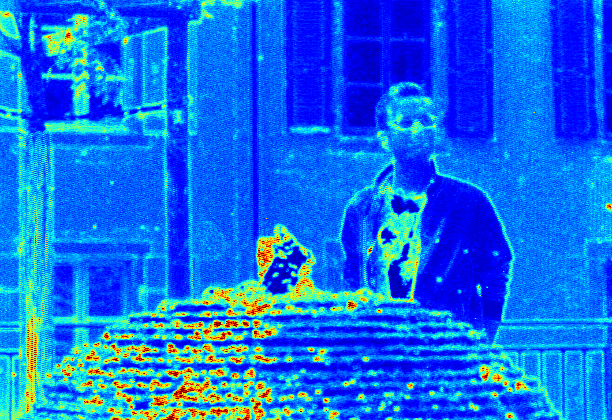} &
\includegraphics[width=0.16\linewidth]{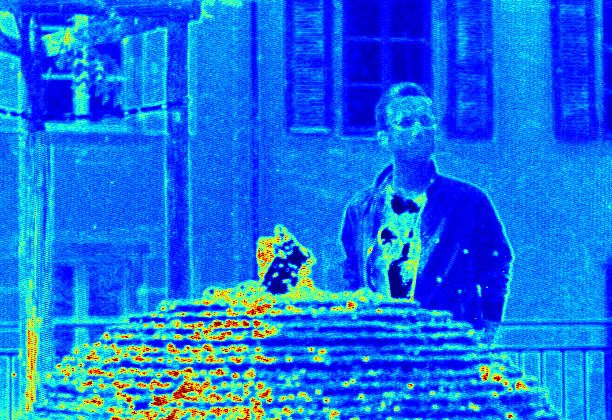} &
\includegraphics[width=0.16\linewidth]{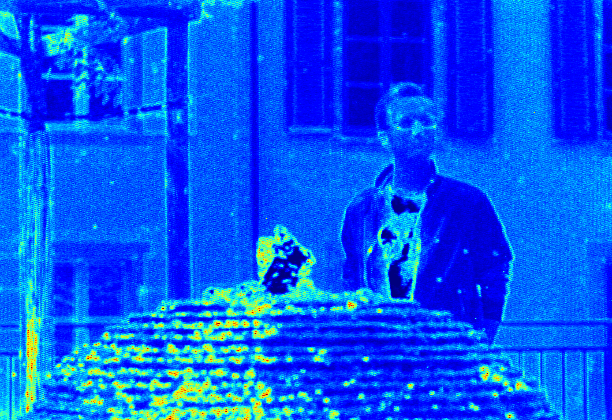}\\
\textbf{\textit{FountainVincent2}}  & 31.21dB-0.25bpp & 30.84dB-0.20bpp & 31.44dB-0.19bpp & 31.93dB-0.21bpp & 31.98dB-0.20bpp & 33.76dB-0.19bpp\\
\includegraphics[width=0.16\linewidth]{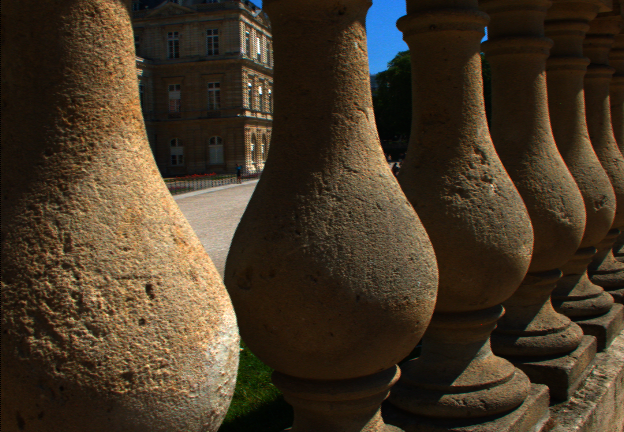} &
\includegraphics[width=0.16\linewidth]{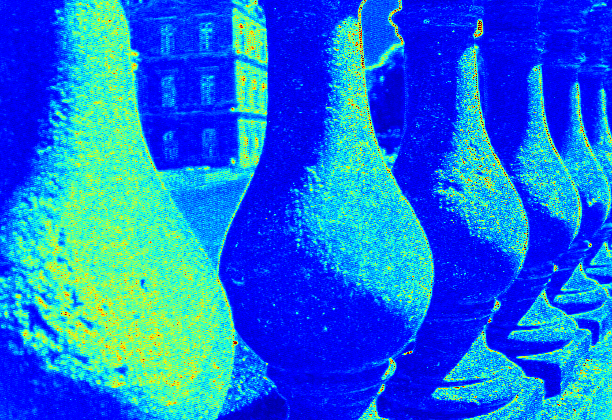} &
\includegraphics[width=0.16\linewidth]{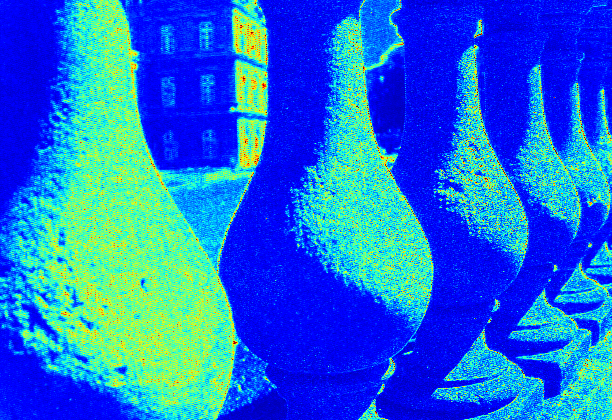} &
\includegraphics[width=0.16\linewidth]{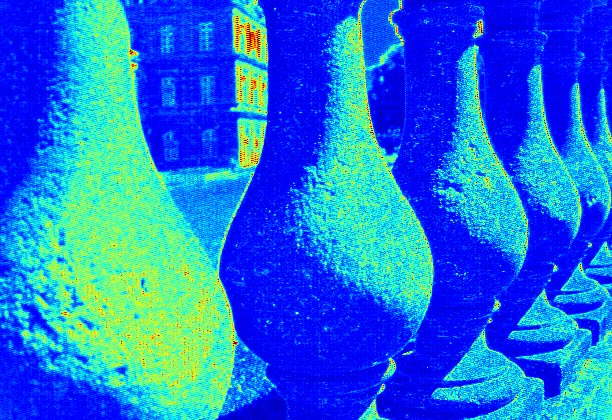} &
\includegraphics[width=0.16\linewidth]{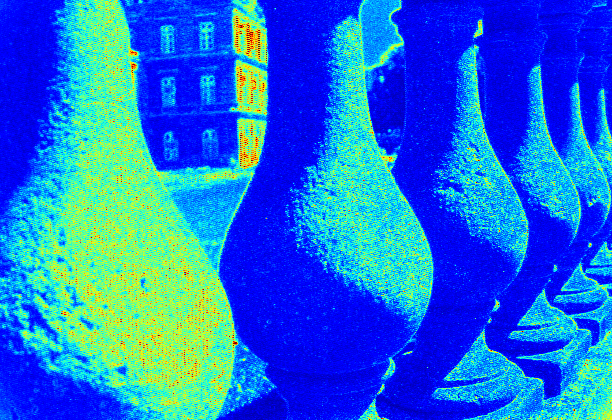} &
\includegraphics[width=0.16\linewidth]{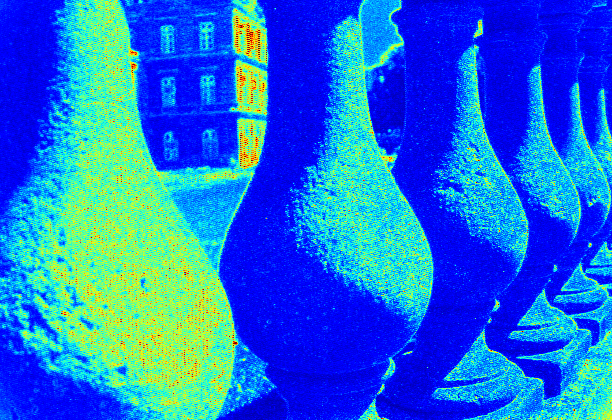} &
\includegraphics[width=0.16\linewidth]{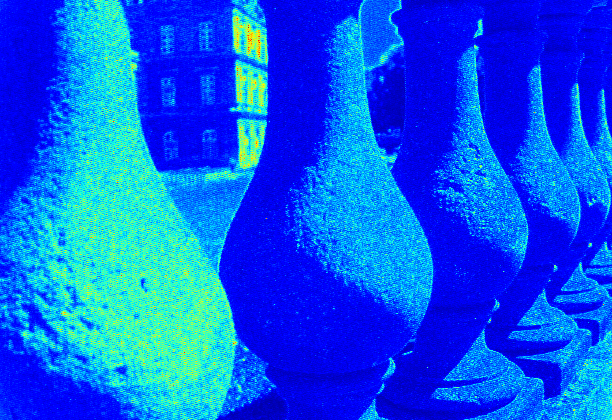}\\
\textbf{\textit{StonePillarsOutside}} & 31.75dB-0.30bpp & 31.79dB-0.19bpp & 31.80dB-0.21bpp & 31.67dB-0.21bpp & 32.25dB-0.21bpp & 33.70dB-0.19bpp\\
\end{tabular}
}
\caption{Averaged error maps of decompressed light fields using different methods, along with the PSNR and bitrate values.}
\label{fig:error_map}
\end{figure*}

\subsection{Compression performance analysis}
\label{sec:comp_perf}
\subsubsection{Rate-distortion}
\label{sec:rate_distortion}
We illustrate in Fig.~\ref{fig:rate_dist} the bit-distortion curves in terms of decoding quality (PSNR) and bitrate (bpp) for all compared methods. We also calculate the BD-PSNR gains using the Bjontegaard metric~\cite{Bjntegaard2001CalculationOA} in Tab.~\ref{table:DB_PSNR} to assess each method and take the results of HEVC-Lozenge as its baseline. We can observe that the proposed method outperforms other methods on most of the scenes by a large margin in both low, medium and high bitrates.  

Although DDLF \cite{jiang2022untrained}, QDLR-NeRF \cite{shi2023light}, and our proposed method all belong to INR-based methods, they exhibit different performances due to their distinct design philosophies. Specifically:
DDLF adopts a design where the transient information is modeled using a GRU module, while the static information is modeled using a deep decoder. However, the GRU architecture performs well only when the light fields have small disparity, as large parallax makes the transition between SAIs hard to be captured by GRU. This explains the performance degradation of DDLF on wide-baseline synthetic light fields.
QDLR-NeRF initially uses an MLP to store the scene information and then employs operations such as low-rank optimization, distillation, and quantization to reduce the model size, ultimately achieving the goal of compression. The quality of the learned scene information directly affects the compression performance. Noise and artifacts that interfere with the learning of scene information can lead to lower compression performance. This is verified by QDLR-NeRF's relatively worse performance on Lytro-captured data.
In comparison, our method stands out due to the cooperation of descriptors and switchable modulators. This feature enables the learning of each SAI to be conducted \textbf{individually}, and detain the impact of factors such as baseline, noise and artifacts. As a result, our method exhibits stable and high performance over two types of light fields.

\subsubsection{Visual comparison}
\begin{figure}
\centering 
\subfigure[DDLF~\cite{jiang2022untrained}]{\label{fig:teaser_DDLF}\includegraphics[width=0.45\linewidth]{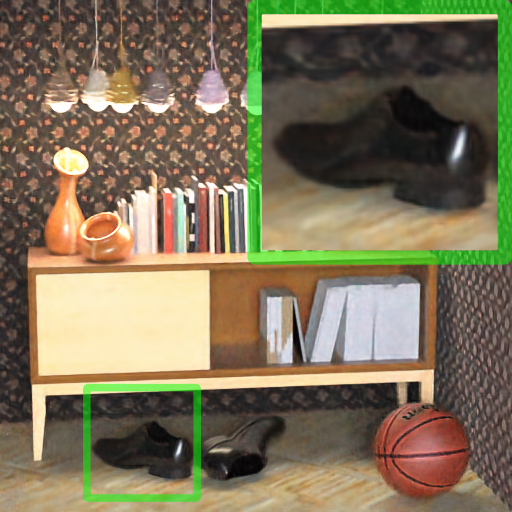}}
\hspace{+0.2cm}
\subfigure[QDLR-NeRF~\cite{shi2023light}]{\label{fig:teaser_QDLR-NeRF}\includegraphics[width=0.45\linewidth]{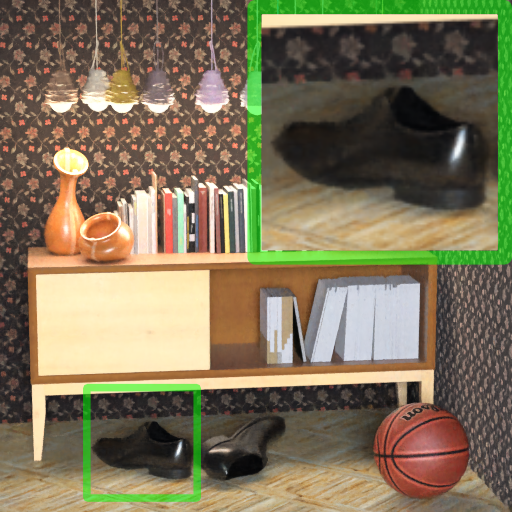}}
\\
\subfigure[Ours]
{\label{fig:teaser_ours}\includegraphics[width=0.45\linewidth]{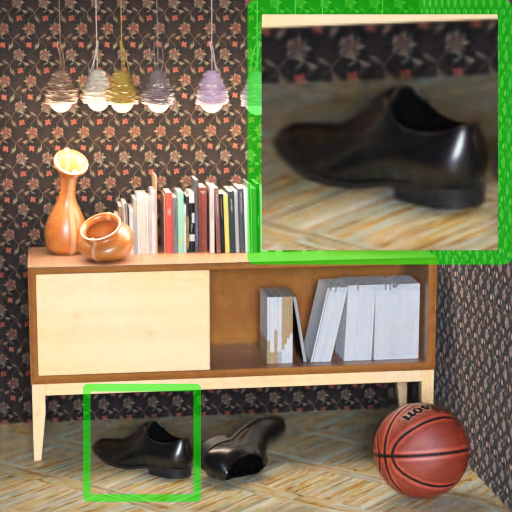}}
\hspace{+0.2cm}
\subfigure[GT View]{\label{fig:teaser_gt}\includegraphics[width=0.45\linewidth]{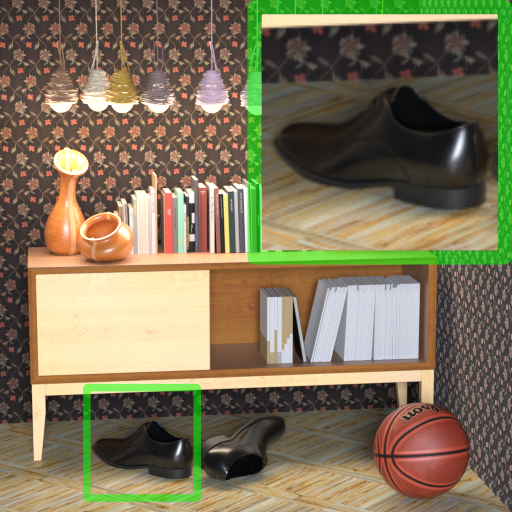}}
\caption{A visualization example of a decoded view using (a). DDLF~\cite{jiang2022untrained} (26.51dB, bpp=0.065), (b). QDLR-NeRF~\cite{shi2023light} (30.25dB, bpp=0.062), (c). Our method (31.60dB, bpp=0.059), (d). Ground-truth view. A local region is zoomed for comparison.}
\label{fig:teaser}
\end{figure}

In Fig.~\ref{fig:error_map}, we present the averaged error maps across all SAIs for each method at a similar bitrate. In the visualization, red indicates a large error value, while blue represents a small error. It's evident that our proposed method outperforms other methods in terms of decoding error, particularly when dealing with highly textured scenes.
Furthermore, Fig.~\ref{fig:teaser} showcases the decoded SAIs of the scene \textit{`sideboard'}, generated by three INR-based methods: DDLF, QDLR-NeRF, and our method. We can notice that our method successfully reconstructs clear floor texture (as seen in the zoomed regions) even at a low bitrate of approximately 0.06 bpp. These error maps and decoded SAIs provide compelling evidence for the effectiveness of our method. 

\subsubsection{Memory consumption and encoding-decoding time}
\label{sec:memory_encoding_decoding}
\begin{figure}[htbp]
    \centering
  \includegraphics[width=\linewidth]{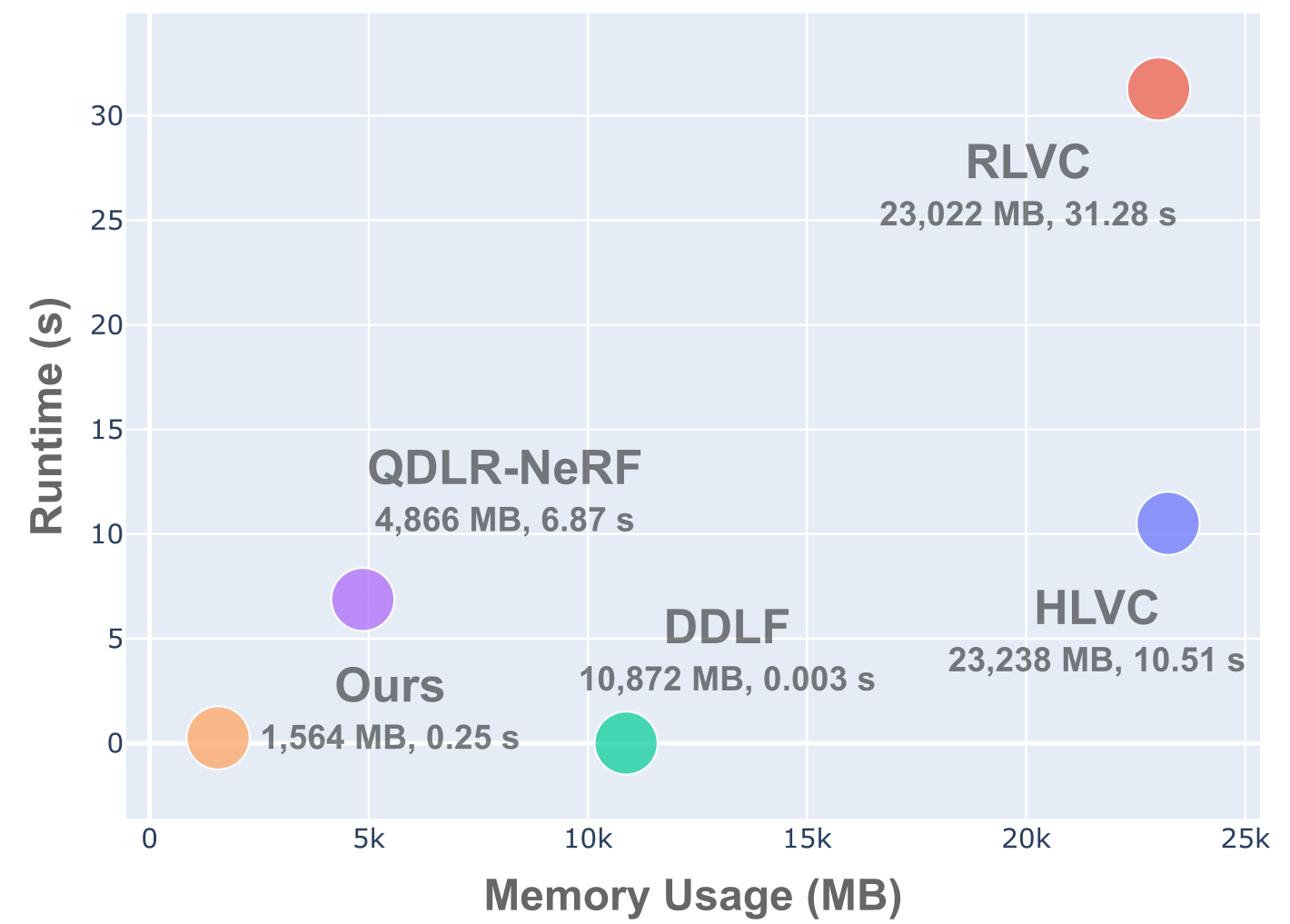}
  \caption{The GPU memory usage and time for decoding per SAI of each learning-based method, measured on Nvidia Titan RTX and the scene `\textit{sideboard}' ($512\times512\times9\times9$).}
  \label{fig:memory_time}
\end{figure}

When evaluating a compression algorithm, both memory consumption and complexity play crucial roles. Lower memory consumption ensures broader hardware support, while decoding time directly impacts the delay in displaying light fields.
In Fig.~\ref{fig:memory_time}, we present the memory usage and decoding time for each learning-based method working on the GPU platform. Among these methods, DDLF\cite{jiang2022untrained} employs a GRU to recurrently process SAIs and decodes a block of views at once, resulting in shorter decoding time but higher memory consumption than ours. The QDLR-NeRF method \cite{shi2023light} adopts a pixel-wise rendering mechanism, leading to a slower decoding procedure. Additionally, due to the complexity of their pipelines, both the HLVC \cite{yang2020Learning} and RLVC \cite{yang2021learning} methods require more memory and time for decoding each SAI.
In contrary, thanks to the switchable modulator design, our network can decode SAI one by one with lower memory consumption, and the fully convolutional network also ensures quick forward pass with less inference time. Though slightly slower than DDLF, our method presents the best trade-off between memory consumption and decoding time.

When considering encoding time, learning-based compression methods have inherent limitations compared to classical compression standards like HEVC and JPEG-Pleno: though methods HLVC~\cite{yang2020Learning} and RLVC~\cite{yang2021learning} exhibit competitive encoding times to HEVC and JPEG-Pleno, they require a large training set for training, and their compression capacity heavily depends on the scale and quality of the training set used. While for INR-based methods, the encoding time mainly consists of supervising the target light field. As a result, their encoding time is generally longer than that of other methods. However, it's noteworthy that in certain applications, encoding time is not as critical as decoding time, as the encoding process can be performed in parallel and offline manners. 

\begin{figure}[htbp]
    \centering
  \includegraphics[width=\linewidth]{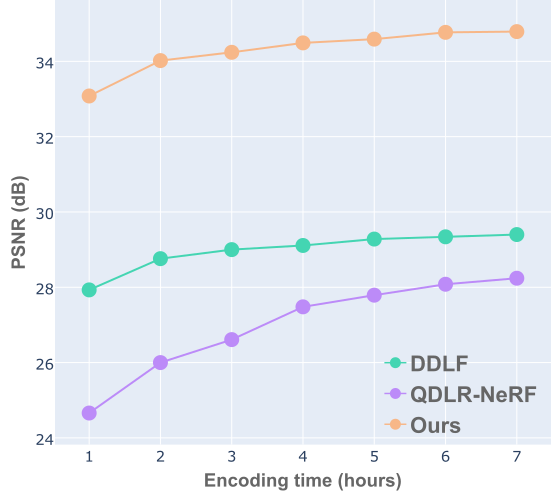}
  \caption{Performance in terms of encoding time for each network-based method. The experiment is carried out on the scene \textit{`sideboard'} ($512\times 512\times 9\times 9$).}
  \label{fig:encoding_time}
\end{figure}

To gain a better understanding of the training efficiency of our method, we provide Fig.~\ref{fig:encoding_time} that illustrates the quality of the decoded light field in relation to training time, and compare the encoding efficiency with other two INR-based methods. We only account for the time spent on the network initialization and exclude the time for low-rank optimization and distillation for the method QDLR-NeRF. We can find in Fig.~\ref{fig:encoding_time} that, even without taking low-rank optimization and distillation into account, the method QDLR-NeRF still needs to gradually improve the performance via a long schedule, while both DDLF and our method can quickly reach a high performance after a short encoding procedure of 1-2 hours, but our method has much higher performance than DDLF. That verifies the encoding efficiency of our method against the other two methods.

\begin{figure*}[!htbp]
    \centering
    \setlength{
    \tabcolsep}{6pt}
    \centering
\resizebox{0.9\textwidth}{!}{
\begin{tabular}{ccccc}
\includegraphics[width=0.18\linewidth]{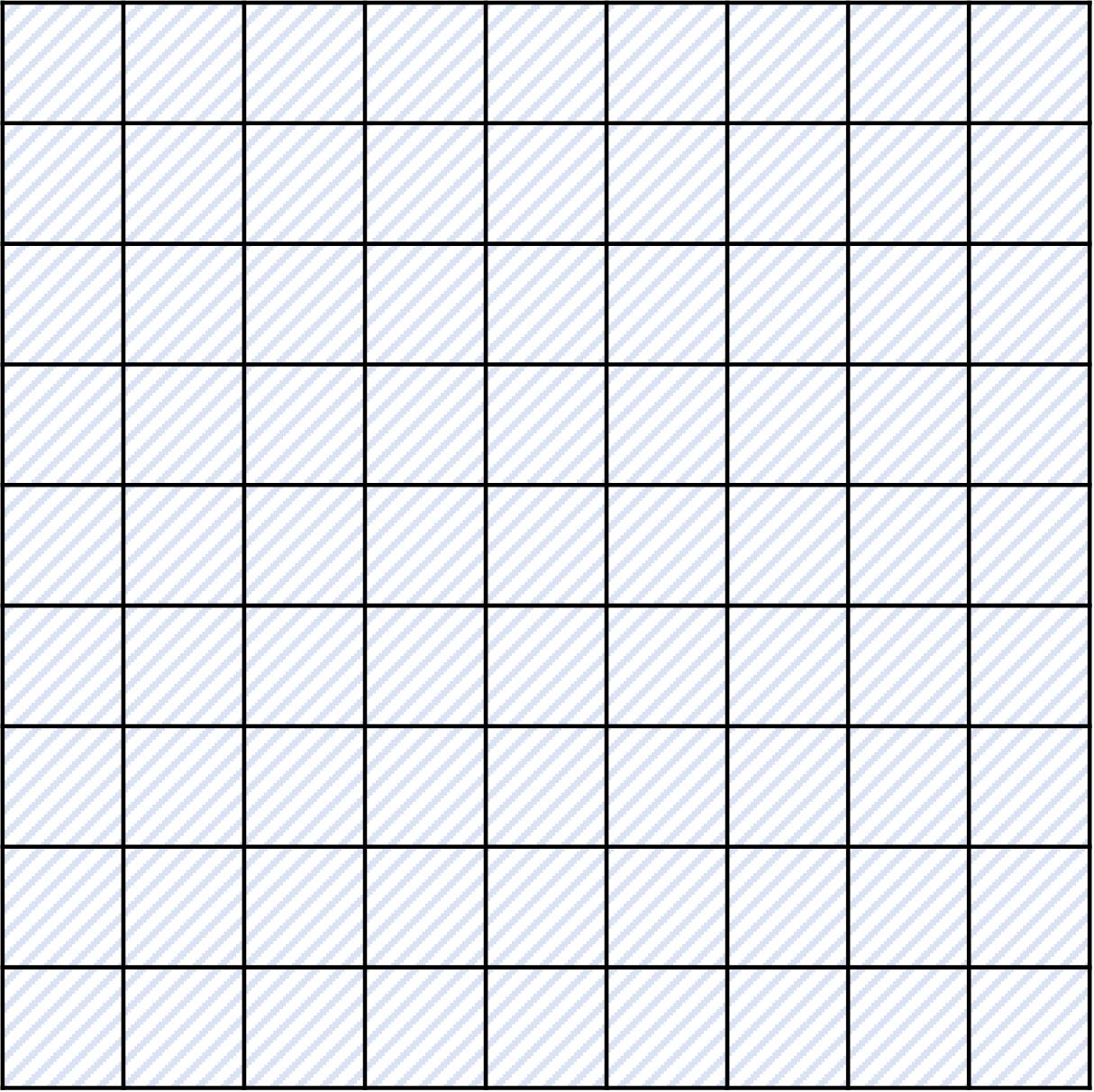} & 
\includegraphics[width=0.18\linewidth]{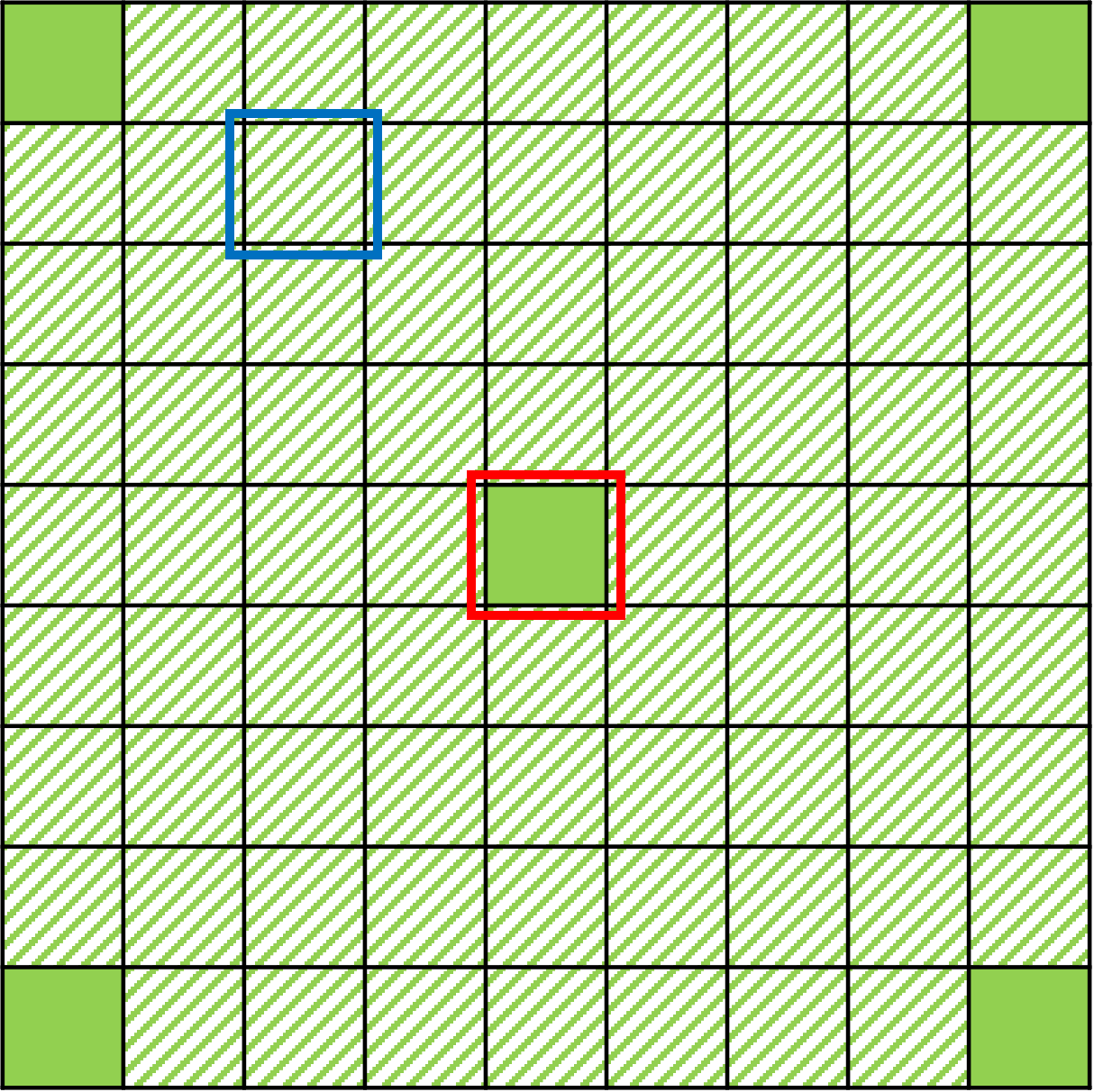} &
\includegraphics[width=0.18\linewidth]{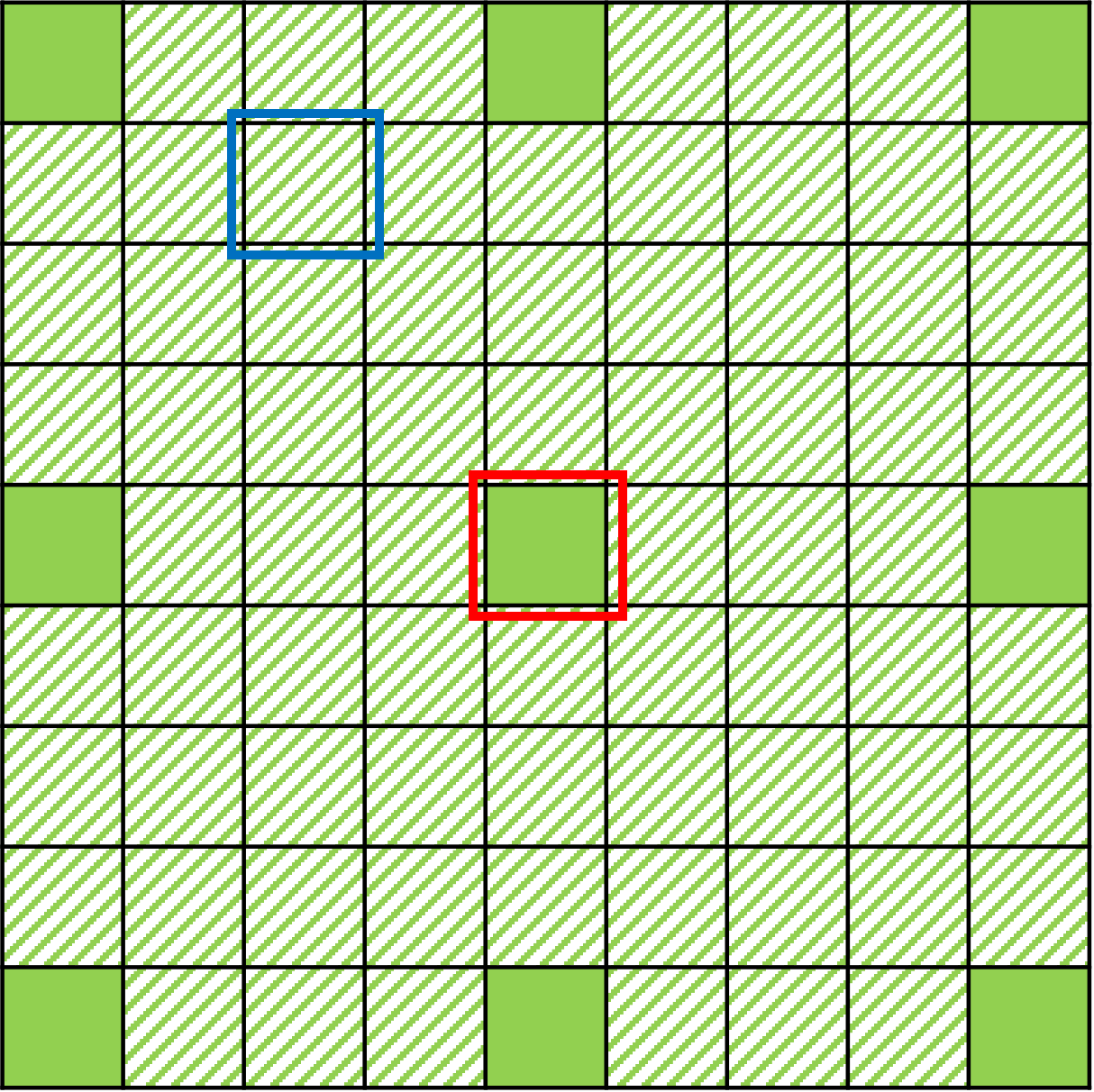} &
\includegraphics[width=0.18\linewidth]{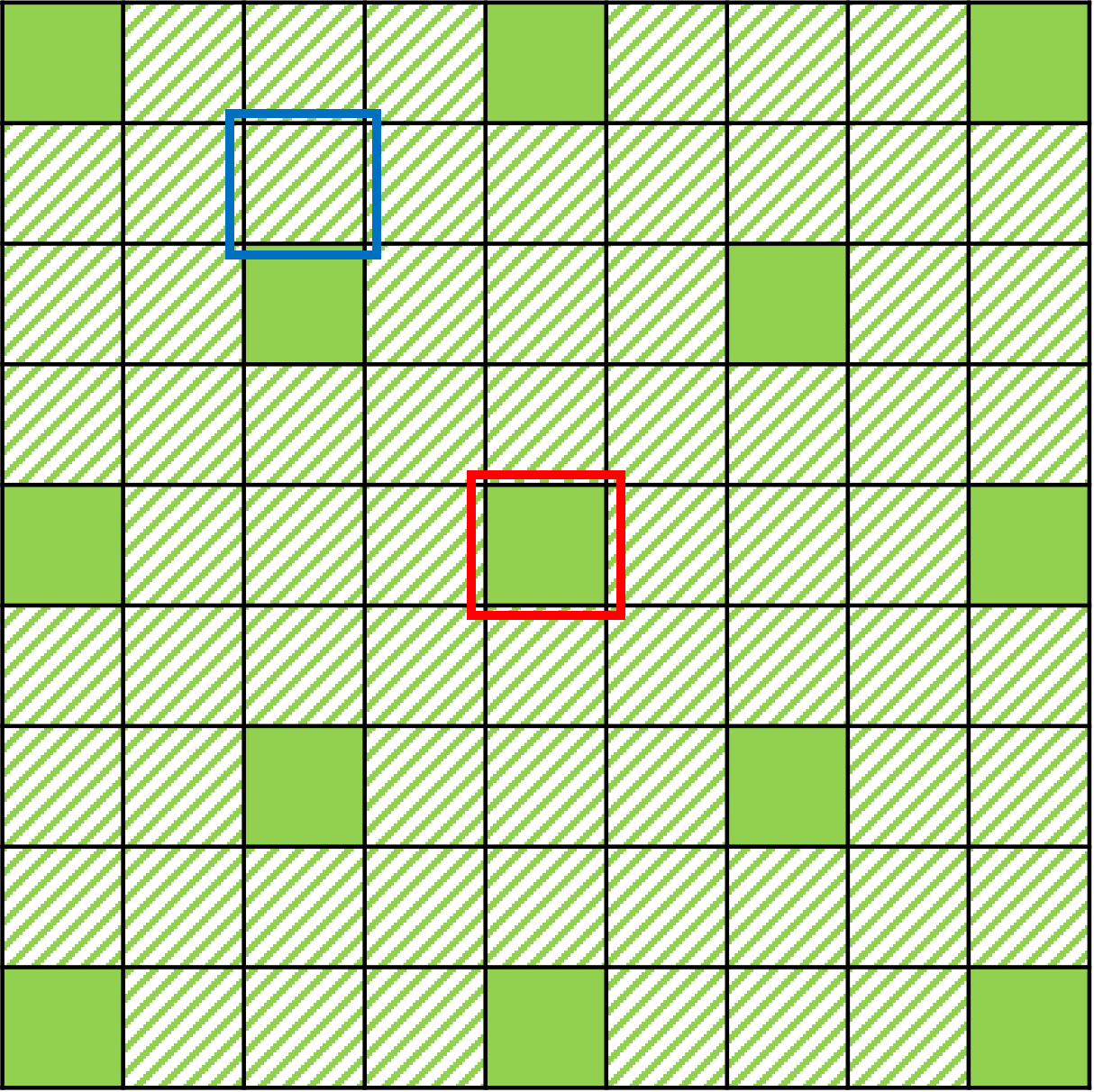} &
\includegraphics[width=0.18\linewidth]{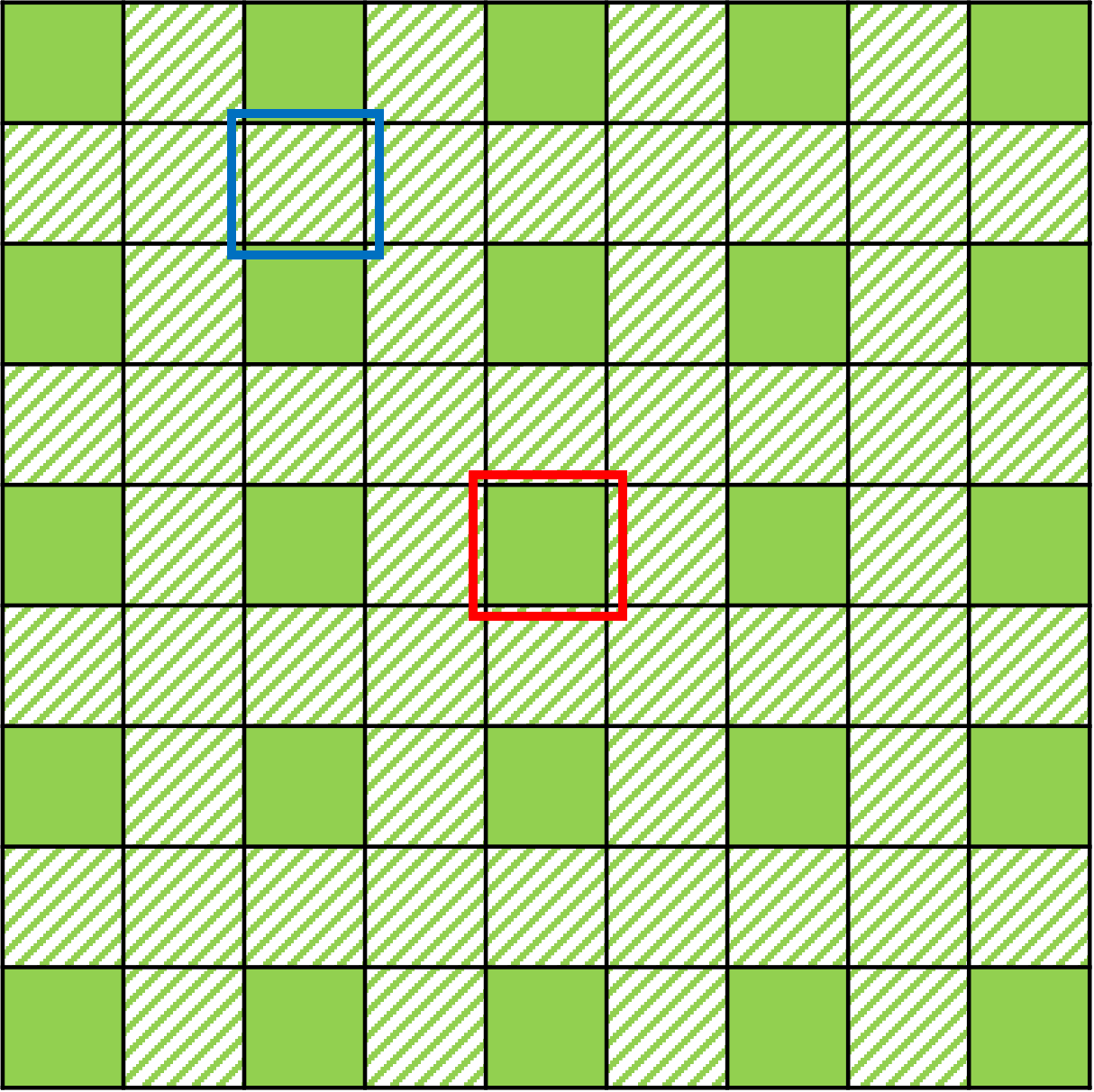}\\
SAI Array & \textcolor{mygreen}{5-SAI Pattern} & \textcolor{mygreen}{9-SAI Pattern} & \textcolor{mygreen}{13-SAI Pattern} & \textcolor{mygreen}{25-SAI Pattern}\\
\includegraphics[width=0.18\linewidth]{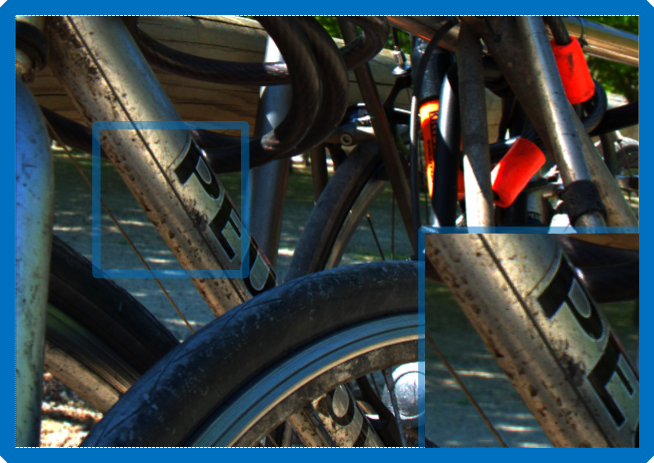} & 
\includegraphics[width=0.18\linewidth]{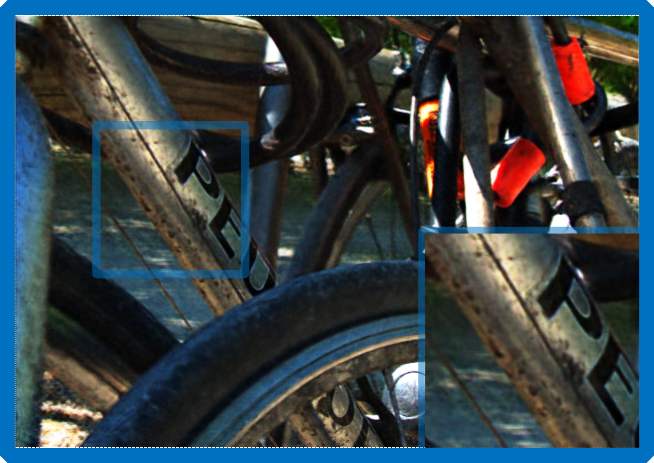} &
\includegraphics[width=0.18\linewidth]{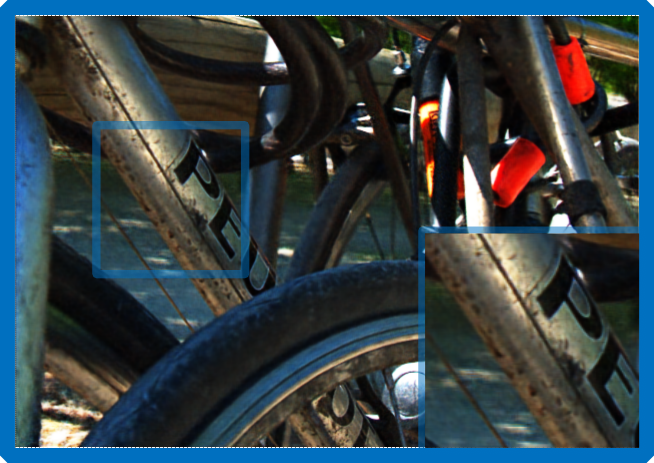} &
\includegraphics[width=0.18\linewidth]{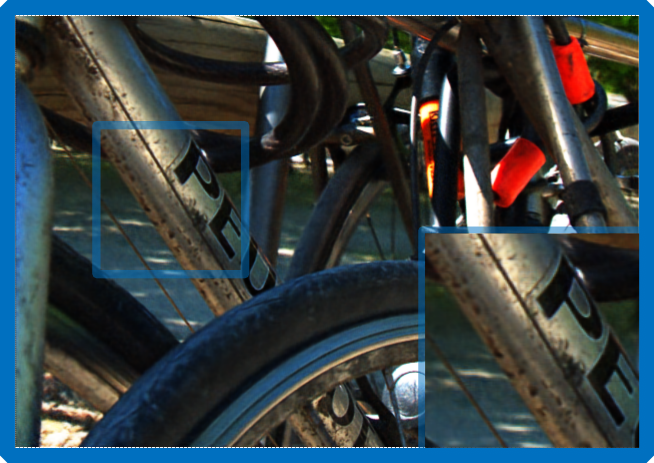} &
\includegraphics[width=0.18\linewidth]{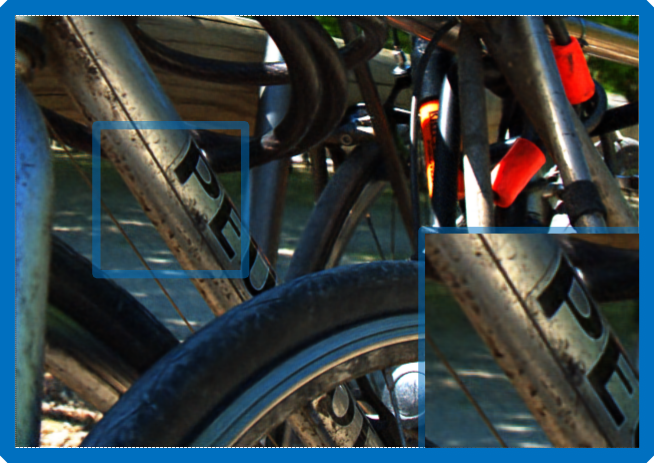}\\
% GT $I_{(2,3)}$ & \textcolor{myblue}{27.01dB} & \textcolor{myblue}{28.67dB} & \textcolor{myblue}{31.28dB} & \textcolor{myblue}{31.68dB}\\
\includegraphics[width=0.18\linewidth]{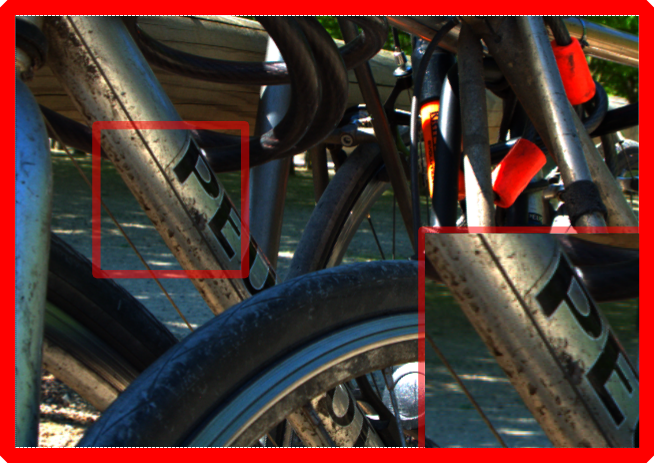} & 
\includegraphics[width=0.18\linewidth]{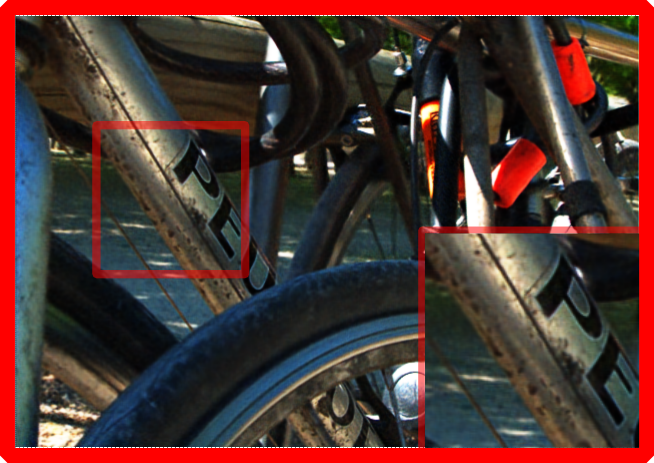} &
\includegraphics[width=0.18\linewidth]{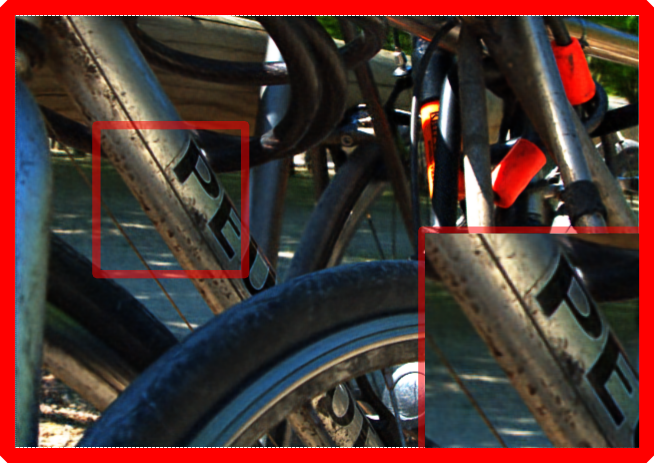} &
\includegraphics[width=0.18\linewidth]{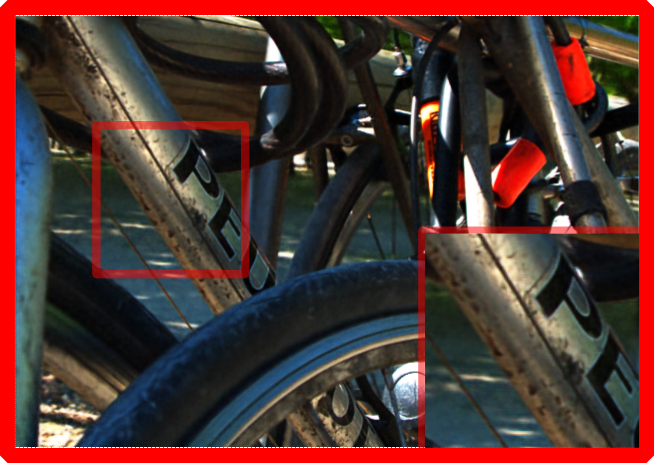} &
\includegraphics[width=0.18\linewidth]{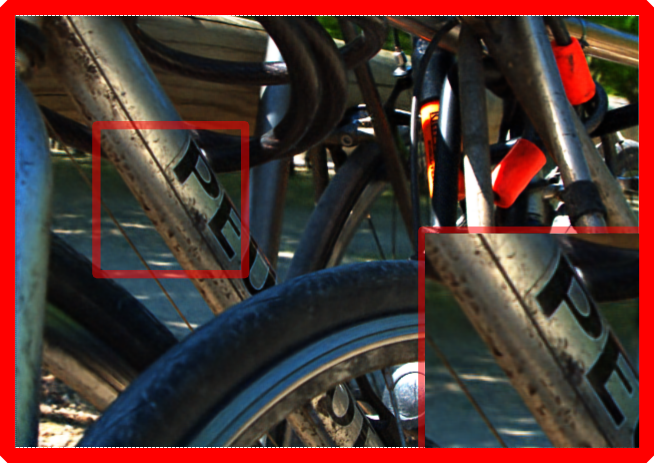}\\
\textcolor{myblue}{GT $I_{(2,3)}$}-\textcolor{red}{$I_{(5,5)}$} & \textcolor{myblue}{27.01dB}-\textcolor{red}{31.52dB} & \textcolor{myblue}{28.67dB}-\textcolor{red}{31.77dB} & \textcolor{myblue}{31.28dB}-\textcolor{red}{31.79dB} & \textcolor{myblue}{31.68dB}-\textcolor{red}{31.92dB}\\
\includegraphics[width=0.18\linewidth]{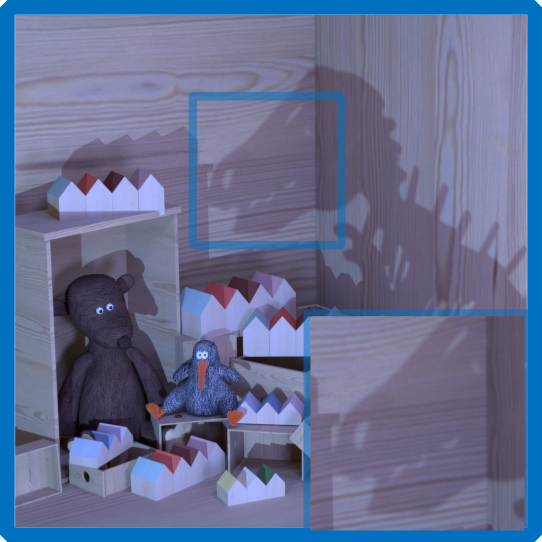} & 
\includegraphics[width=0.18\linewidth]{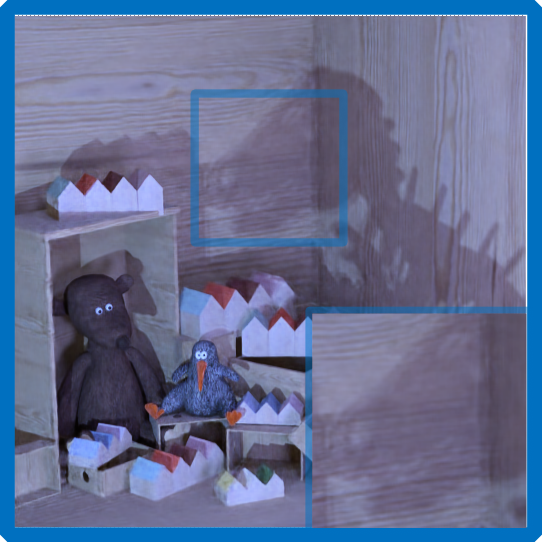} &
\includegraphics[width=0.18\linewidth]{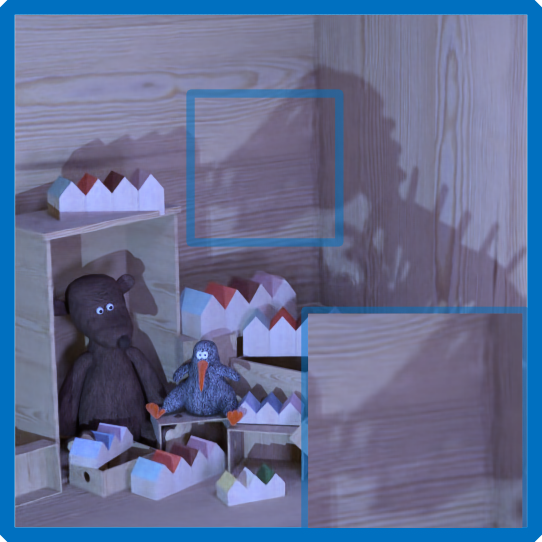} &
\includegraphics[width=0.18\linewidth]{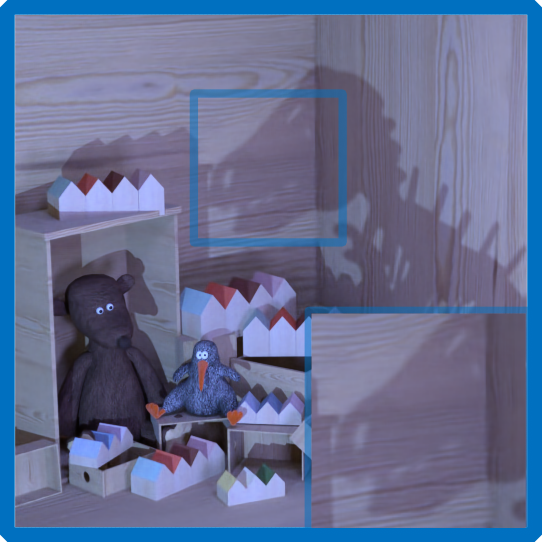} &
\includegraphics[width=0.18\linewidth]{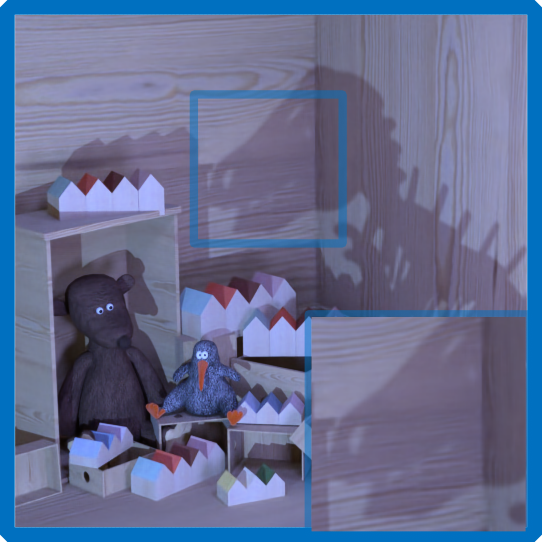}\\
% \textcolor{myblue}{GT $I_{(2,3)}$} & \textcolor{myblue}{32.25dB} & \textcolor{myblue}{33.89dB} & \textcolor{myblue}{38.60dB} & \textcolor{myblue}{39.19dB}\\
\includegraphics[width=0.18\linewidth]{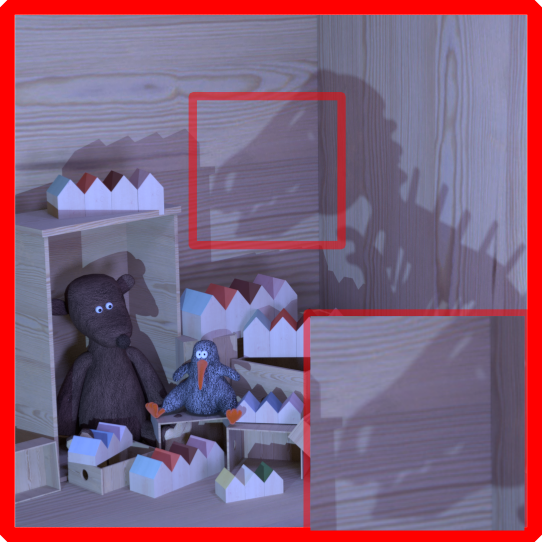} & 
\includegraphics[width=0.18\linewidth]{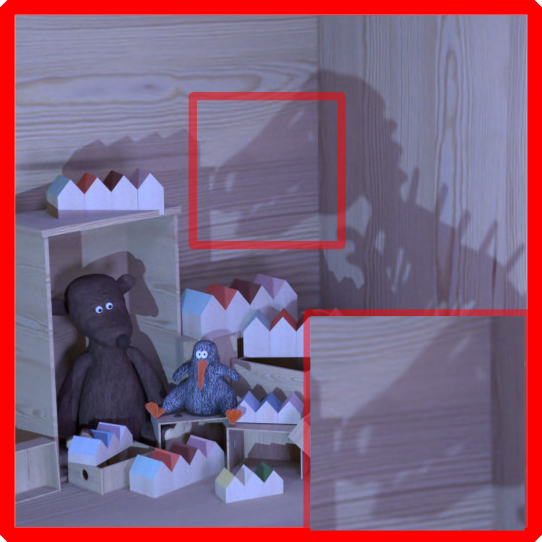} &
\includegraphics[width=0.18\linewidth]{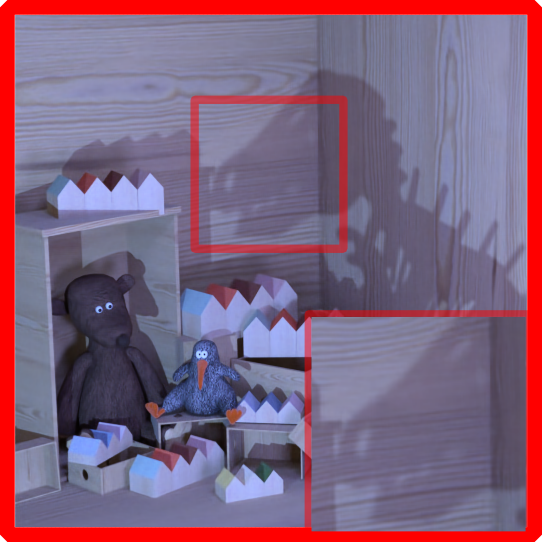} &
\includegraphics[width=0.18\linewidth]{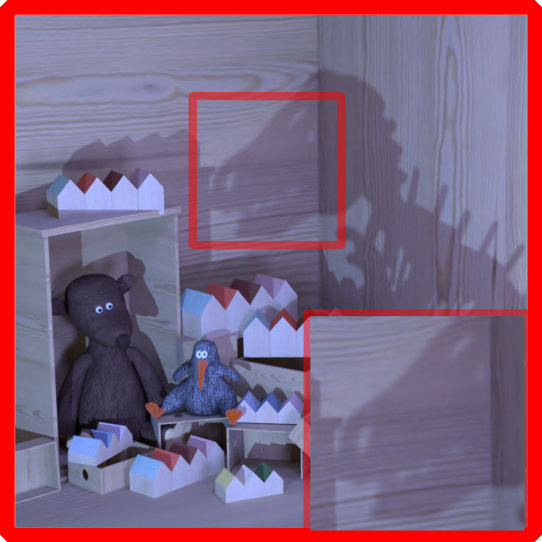} &
\includegraphics[width=0.18\linewidth]{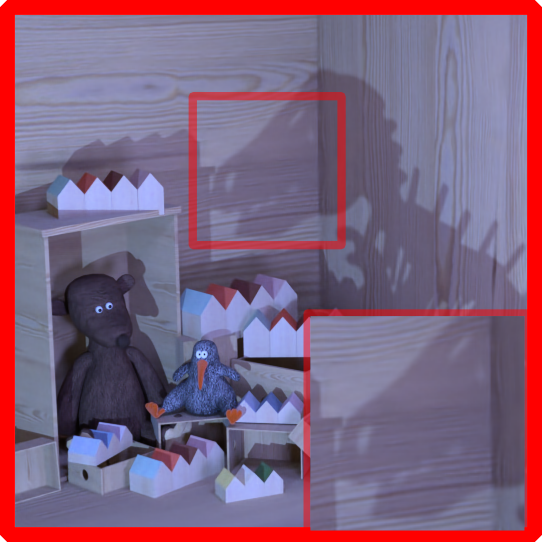}\\
\textcolor{myblue}{GT $I_{(2,3)}$}-\textcolor{red}{$I_{(5,5)}$} & \textcolor{myblue}{32.25dB}-\textcolor{red}{39.65dB} & \textcolor{myblue}{33.89dB}-\textcolor{red}{40.00dB} & \textcolor{myblue}{38.60dB}-\textcolor{red}{40.09dB} & \textcolor{myblue}{39.19dB}-\textcolor{red}{40.20dB}\\
\end{tabular}
}
\caption{Several sebset patterns for retraining and corresponding rendered views. The first row show 4 subset patterns with increasing number of SAIs for retraining. We also illustrate the rendered views $\hat{I}_{(5,5)}$ and $\hat{I}_{(2,3)}$ that use the involved/uninvolved modulators in the (third and fifth)/(second and fourth) rows. The second and the third rows show rendered views for \textit{`bikes'} and the fourth and the fifth rows show results for \textit{`dino'}.}
\label{fig:modulator_patterns}
\end{figure*}

\subsection{Transfer of modulators}
\label{sec:AIT}
We defined two types of kernels in our network design: descriptors, which store scene information, and modulators, which control the rendering of SAIs with respect to the desired perspectives. The experiment in this section demonstrates that the modulators can be non-scene-specific if the descriptors are appropriately aligned, i.e. the modulators learned on one light field can be transferred to the new light fields.
More precisely, we take two light fields $L_{1}$ and $L_{2}$ each with $9\times 9$ SAIs as an example, and carry out the following steps: 
\begin{itemize}
\item[1.] \textbf{\textit{Pretraining on $L_{1}$}}: we first train the network using all SAIs of $L_{1}$, during which both descriptors $\{K^{d}_{L_{1}}\}$ and $9+9$ sets of modulators $\{K^{m}_{u,L_{1}},K^{m}_{v,L_{1}},b^{m}_{u,L_{1}},b^{m}_{v,L_{1}}\}$ are updated. 
\item[2.] \textbf{\textit{Retraining descriptors on $L_{2}$}}: we then fix the learned modulators $\{K^{m}_{u,L_{1}},K^{m}_{v,L_{1}},b^{m}_{u,L_{1}},b^{m}_{v,L_{1}}\}$ and retrain descriptors using a SUBSET of SAIs (e.g. a sparse $3\times 3$ views) of $L_{2}$ for one epoch to obtain $\{K^{d}_{L_{2}}\}$.  
\item[3.] \textbf{\textit{Rendering all SAIs of $L_{2}$}:} we render all SAIs of $L_{2}$ with the updated descriptors $\{K^{d}_{L_{2}}\}$ and modulators $\{K^{m}_{u,L_{1}},K^{m}_{v,L_{1}},b^{m}_{u,L_{1}},b^{m}_{v,L_{1}}\}$.
\end{itemize}

In our experiment, we specifically utilize a subset of SAIs from $L_{2}$ to retrain the descriptors for step 2, indicating that only a part of modulators are involved in this procedure. There are two main reasons for adopting sparse sampling instead of all views:
(a). The SAIs inside the subset provide information about the new scene. Retraining on these views helps align the descriptors for storing new scene information.
(b). It is important to note that the modulators for the views outside the subset are entirely excluded from the retraining process. If these modulators can successfully work with the descriptors to synthesize SAIs, it suggests that the modulators are non-scene-specific and its functionality of modulation is transferrable. Conversely, if the excluded modulators fail to generate SAIs while those involved in the retraining procedure perform well for rendering, it would imply that the functionalities of the modulator and descriptor are scene-specific and are endowed only by training on the current scene. We test two cases with several subset patterns: 
\begin{itemize}
\item[(a).] Pretraining the network on the scene \textit{`danger'} and retraining descriptors on \textit{`bikes'}.
\item[(b).] Pretraining the network on the scene \textit{`boxes'} then retraining descriptors on the scene \textit{`dino'}.
\end{itemize}
As both \textit{`danger'} and \textit{`bikes'} are captured using the same Lytro camera, while \textit{`boxes'} and \textit{`dino'} are synthesized using different camera array configurations, These two cases respectively represent the transfer of modulators between cameras with the same and distinct configurations. 
Fig.~\ref{fig:modulator_patterns} showcases the subset patterns and rendered SAIs. The first row depicts the subset patterns with an increasing number of SAIs used for retraining step, where the views inside the subset are colored in green, and other excluded views are noted with green slashes. The checks framed with red and blue boxes are positions of the SAIs to be shown from the second to the fifth rows, they respectively represent SAIs rendered using modulators involved and not involved (noted as `\textbf{involved modulator}' and `\textbf{uninvolved modulator}') in the retraining procedure. Rows two and three display generated SAIs of \textit{`bikes'} and row four and five exhibit rendered SAIs of the scene \textit{`dino'}.

We observe that both the involved and uninvolved modulators can work with the descriptors to generate SAIs of the new light fields, even this transfer occurs between cameras with different configurations. More SAIs in the subset improves the quality of views rendered with uninvolved modulators. This is because using more SAIs in the retraining will better align the descriptors with the modulators. Furthermore, SAIs generated using involved modulators show better quality than those generated using uninvolved modulators, as modulators involved in the retraining step always better match descriptors than those uninvolved ones. Let us note that such a modulator transfer operation also implies a new solution for view synthesis task, one can generate novel dense views by transferring the learned modulators to the target light field.

\section{Ablation Study}
\subsection{Proportion of modulator parameter}
\label{sec:sk_vs_ak}
\begin{figure}[!thb]
    \centering
  \includegraphics[width=0.99\linewidth]{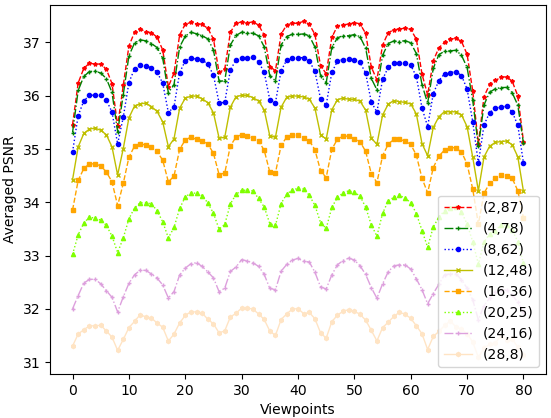}
  \caption{The averaged PSNR in terms of viewpoints, with the index of the top left corner view being labeled as `0' and the index of the bottom right corner view as `80'.}
  \label{fig:psnr_view}
\end{figure}

When adopting our network architecture for light field compression, the proportion of modulator parameters plays a key role in determining the compression performance. To explore the optimal proportion of modulators for network design, we conducted experiments by varying the proportions of modulator parameters under a fixed total parameter constraint. Tab.~\ref{table:proportion} illustrated the average PSNR and quality variance across $9\times 9$ views for 8 different scenes, where $c_{m}$ and $c_{d}$ respectively denote the channel numbers for modulator and descriptor. To give a better insight on performance variation in terms of viewpoints, we also display in Fig.~\ref{fig:psnr_view} the averaged PSNR on 8 scenes for different viewpoints.

From both Fig.~\ref{fig:psnr_view} and Tab.~\ref{table:proportion}, we can observe that the proportion of modulator parameters directly affects the network's performance. Under similar total parameter number constraint, higher proportion of modulator parameters means lower proportion for descriptors. It results in relatively lower averaged PSNR and smaller quality variance among views. This occurs because the network has a limited number of parameter for storing scene information, but enough parameters to modulate the rendering of SAIs. Instead, reducing the proportion of modulator parameter will spare more parameters for descriptors, which improves the quality of the decoded views, but weakens network's modulation capability and leads to larger quality variance. 
The above observation can serve as a guideline for network design for different compression demands: when aiming for a high-quality representation of the entire light field, it is preferable to use a lower proportion of modulator parameters. However, if maintaining consistency between SAIs is a priority, a higher proportion is recommended.

\begin{table}[!th]
\begin{center}
\caption{The averaged PSNR and variance in terms of different proportions of modulator parameter, calculated over 81 views of 8 scenes under a fixed total parameter number constraint.}
\label{table:proportion}
\begin{tabular}{c|c|c|c}
\hline
\hline
$(c_{m},c_{d})$ & Proportion & PSNR(dB) & Variance \\
\hline
(28,8)        & 97\%       & 31.67      & 0.085       \\
\hline
(24,16)        & 93\%       & 32.54      & 0.102       \\
\hline
(20,25)        & 88\%       & 33.77      & 0.147       \\
\hline
(16,36)         & 80\%       & 34.80      & 0.186       \\
\hline
(12,48)         & 70\%       & 35.51      & 0.228       \\
\hline
(8,62)         & 55\%       & 36.18      & 0.275       \\
\hline
(4,78)         & 33\%       & 36.62      & 0.312       \\
\hline
(2,87)         & 19\%       & 36.81      & 0.345       \\
\hline
\hline
\end{tabular}
\end{center}
\end{table}

\subsection{Effectiveness of kernel design}
\label{sec:kernel_design}
To validate our proposals of modulator allocation and kernel tensor decomposition, under the constraint of similar total number of network parameters, we tested three network variants:

\begin{itemize}
\item[(a).] The network without both modulator allocation and kernel tensor decomposition designs, which is denoted as \textit{Net}$\dag$.
\item[(b).] The network without modulator allocation but with kernel tensor decomposition, this one is denoted as \textit{Net*}.
\item[(c).] The network that adopts both modulator allocation and kernel tensor decomposition designs, this variant is denoted as \textit{Net}.
\end{itemize}

We measure the averaged PSNR of the networks having small, moderate, and large parameter numbers, corresponding to low, intermediate, and high bitrates in the context of compression. Tab.~\ref{tab_ablation2} summarizes the performance of each network variant for different numbers of parameters. 
The application of modulator allocation results in significant parameter savings that can be allocated to the descriptors for performance enhancement.
And the adoption of tensor decomposition enables the reduction of the number of parameters in kernels, thereby accommodating more kernels in the network. The combination of both modulator allocation and kernel tensor decomposition results in a notable improvement of the network's performance.

\setlength{\tabcolsep}{4pt}
\begin{table}[t!]
\begin{center}
\caption{Performance comparison between different network variants for low, moderate and high numbers of parameters. The best performances are in bold.}
\label{tab_ablation2}
\begin{tabular}{c|c|ccc}
\hline
\hline
 & \#Param &  103K & 325K & 809K \\
\hline
\multirow{2}*{\textit{Net}$\dag$} & $(c_{m},c_{d})$ & (2,11) &  (2,34)  &  (2,72) \\
   & PSNR (dB)  & 27.43   & 33.21    & 37.16   \\
\hline
\multirow{2}*{\textit{Net*}} & $(c_{m},c_{d})$ & (2,16)  & (2,47)  & (2,97)  \\
& PSNR (dB)  & 28.78   & 34.16  & 37.92   \\
\hline
\multirow{2}*{\textit{Net}} & $(c_{m},c_{d})$ & (2,48)  & (2,93)  & (2,153)  \\
& PSNR (dB)  & \textbf{33.95}   & \textbf{37.47}  & \textbf{39.98}   \\
\hline
\hline
\end{tabular}
\end{center}
\end{table}

\subsection{Contributions of each network design}
\label{sec:contributions}
To highlight the contribution of each network design step, we evaluated the performance evolution after implementing each design step (including modulator allocation, kernel tensor decomposition, and quantization). In Tab.~\ref{table:building_block}, we present the average PSNR results for eight tested scenes and the corresponding network size when applying each design step. For comparison, we consider the original network configuration without modulator allocation, tensor decomposition, and quantization as the baseline, it has channel numbers $(c_m,c_d)=(2,48)$. Due to the high angular resolution $(U,V)=(9,9)$, when without adopting modulator allocation, even if $c_m=2$ is much smaller than $c_d=48$, the network still has a large proportion of parameter allocated to modulators.
Therefore we can observe about $3\times$ compression from 100\% to 31.1\% when applying modulator allocation technique with only 0.15dB performance degradation. Around 0.6dB loss is caused by the tensor decomposition technique, please note that tensor decomposition is a typical network compression method, other advanced decomposition method is likewise applicable to our method. Finally, the quantization operation brings 0.8dB degradation after compacting the network size from 20.93\% to 9.86\%. These techniques globally realize more than $10\times$ compression with about 1.6dB quality degradation. Users can select techniques to be used according to their desired decoding quality and model size for the compression task. 

\begin{table}[t]
\begin{center}
\caption{PSNR averaged on the $8$ test light fields at the different steps of the proposed workflow: Original network (Net-org), network gradually after using modulator allocation (Net-ma), after using tensor decomposition (Net-td) and after quantization (Net-qt)).}
\label{table:building_block}
\begin{tabular}{c|c|c|c|c}
\hline
\hline
Metrics & Net-org & Net-ma & Net-td & Net-qt \\
\hhline{-----}
\hline
PSNR & 34.77dB & 34.62dB & 33.95dB & 33.13dB \\
\hhline{-----}
Size & 100\% & 31.10\% & 20.93\% & 9.86\% \\
\hline
\hline
\end{tabular}
\end{center}
\end{table}

\section{Conclusion}
In this paper, we address the challenge of light field compression by proposing a novel compact neural representation. Our method utilizes two types of complementary kernels: descriptors and modulators. Descriptors capture scene information, while modulators serve to modulate the rendering of different SAIs.
To enhance the network's compactness, we propose allocating modulators across two angular dimensions and decomposing the kernel tensor into low-dimensional components. Through extensive experiments, we demonstrate that our network-based representation outperforms other compression methods while consuming less computational resources.
Furthermore, we highlight that the modulators exhibit a non-scene-specific nature and can be transferred to new light field data for rendering dense views. This finding suggests a new approach to view synthesis methods, introducing a distinct philosophy in this field.

{\small
\bibliographystyle{unsrt}
\bibliography{egbib,reference}
}

\vfill

\end{document}